# Title page

**Title**: AI in radiological imaging of soft-tissue and bone tumours: a systematic review evaluating against CLAIM and FUTURE-AI guidelines

**Authors**: Douwe J. Spaanderman MSc[1*], Matthew Marzetti MSc[2,3*], Xinyi Wan MSc[1*], Andrew F. Scarsbrook MD[4,5], Philip Robinson MD[4], Edwin H.G. Oei MD, PhD[1], Jacob J. Visser MD, PhD[1], Robert Hemke MD, PhD[6], Kirsten van Langevelde MD, PhD[7], David F. Hanff MD[1], Geert J.L.H. van Leenders MD, PhD[8], Cornelis Verhoef MD, PhD[9], Dirk J. Grünhagen MD, PhD[9], Wiro J. Niessen PhD[1,10], Stefan Klein PhD[1,†], Martijn P.A. Starmans PhD[1,8,†]

*Shared first author
†Shared last author

**Affiliations**:
1. Department of Radiology and Nuclear Medicine, Erasmus MC Cancer Institute, University Medical Center Rotterdam, Rotterdam, the Netherlands
2. Department of Medical Physics, Leeds Teaching Hospitals NHS Trust, UK
3. Leeds Biomedical Research Centre, University of Leeds, UK
4. Department of Radiology, Leeds Teaching Hospitals NHS Trust, UK
5. Leeds Institute of Medical Research, University of Leeds, UK
6. Department of Radiology and Nuclear Medicine, Amsterdam UMC, Amsterdam, the Netherlands
7. Department of Radiology, Leiden University Medical Center, Leiden, the Netherlands
8. Department of Pathology, Erasmus MC Cancer Institute, University Medical Center Rotterdam, Rotterdam, the Netherlands
9. Department of Surgical Oncology, Erasmus MC Cancer Institute, University Medical Center Rotterdam, Rotterdam, the Netherlands
10. Faculty of Medical Sciences, University of Groningen, Groningen, the Netherlands

**Corresponding author**
Douwe J. Spaanderman
Internal postal address: P.O. Box 2040, 3000 CA Rotterdam, the Netherlands, Na-2624
Visiting address: office Na-2624, Dr. Molewaterplein 40, 3015 GD Rotterdam, the Netherlands
d.spaanderman@erasmusmc.nl
+31-10-7041026

**Authors addresses**
Douwe J. Spaanderman

Department of Radiology and Nuclear Medicine, Erasmus MC Cancer Institute, University Medical Center Rotterdam, Dr. Molewaterplein 40, 3015 GD Rotterdam, the Netherlands

Matthew Marzetti

Department of Medical Physics, Level 1, Bexley Wing, St James's University Hospital, Beckett Street, Leeds, West Yorkshire, LS9 7TF, United Kingdom

Xinyi Wan

Department of Radiology and Nuclear Medicine, Erasmus MC Cancer Institute, University Medical Center Rotterdam, Dr. Molewaterplein 40, 3015 GD Rotterdam, the Netherlands

Andrew F. Scarsbrook

Department of Nuclear Medicine, Level 1, Bexley Wing, St James's University Hospital, Beckett Street, Leeds, West Yorkshire, LS9 7TF, United Kingdom

Philip Robinson

Leeds Biomedical Research Centre, University of Leeds, Chapel Allerton Hospital, Chapeltown Road, Leeds, LS7 4SA, United Kingdom

Edwin H.G. Oei

Department of Radiology and Nuclear Medicine, Erasmus MC Cancer Institute, University Medical Center Rotterdam, Dr. Molewaterplein 40, 3015 GD Rotterdam, the Netherlands

Jacob J. Visser

Department of Radiology and Nuclear Medicine, Erasmus MC Cancer Institute, University Medical Center Rotterdam, Dr. Molewaterplein 40, 3015 GD Rotterdam, the Netherlands

Robert Hemke

Department of Radiology and Nuclear Medicine, Amsterdam University Medical Center Meibergdreef 9, 1105AZ Amsterdam, the Netherlands

Kirsten van Langevelde

Department of Radiology, Leiden University Medical Center
Albinusdreef 2, 2333 ZA Leiden, the Netherlands

David F. Hanff

Department of Radiology and Nuclear Medicine, Erasmus MC Cancer Institute, University Medical Center Rotterdam, Dr. Molewaterplein 40, 3015 GD Rotterdam, the Netherlands

Geert J.L.H. van Leenders
Department of Pathology, Erasmus MC Cancer Institute, University Medical Center Rotterdam,
Dr. Molewaterplein 40, 3015 GD Rotterdam, the Netherlands

Cornelis Verhoef
Department of Surgical Oncology, Erasmus MC Cancer Institute, University Medical Center Rotterdam,
Dr. Molewaterplein 40, 3015 GD Rotterdam, the Netherlands

Dirk J. Grünhagen
Department of Surgical Oncology, Erasmus MC Cancer Institute, University Medical Center Rotterdam, Dr. Molewaterplein 40, 3015 GD Rotterdam, the Netherlands

Wiro J. Niessen
Faculty of Medical Sciences, University of Groningen
Antonius Deusinglaan 1, 9713 AV Groningen, the Netherlands

Martijn P.A. Starmans
Department of Radiology and Nuclear Medicine, Erasmus MC Cancer Institute, University Medical Center Rotterdam, Dr. Molewaterplein 40, 3015 GD Rotterdam, the Netherlands

Stefan Klein
Department of Radiology and Nuclear Medicine, Erasmus MC Cancer Institute, University Medical Center Rotterdam, Dr. Molewaterplein 40, 3015 GD Rotterdam, the Netherlands

**Title:**

AI in radiological imaging of soft-tissue and bone tumours: a systematic review evaluating against CLAIM and FUTURE-AI guidelines

**Summary**

Background:

Soft-tissue and bone tumours (STBT) are rare, diagnostically challenging lesions with variable clinical behaviours and treatment approaches. This systematic review aims to provide an overview of Artificial Intelligence (AI) methods using radiological imaging for diagnosis and prognosis of these tumours, highlighting challenges in clinical translation, and evaluating study alignment with the Checklist for AI in Medical Imaging (CLAIM) and the FUTURE-AI international consensus guidelines for trustworthy and deployable AI to promote the clinical translation of AI methods.

Methods:

The systematic review identified literature from several bibliographic databases, covering papers published before 17/07/2024. Original research published in peer-reviewed journals, focused on radiology-based AI for diagnosis or prognosis of primary STBT was included. Exclusion criteria were animal, cadaveric, or laboratory studies, and non-English papers. Abstracts were screened by two of three independent reviewers to determine eligibility. Included papers were assessed against the two guidelines by one of three independent reviewers. The review protocol was registered with PROSPERO (CRD42023467970).

Findings:

The search identified 15,015 abstracts, from which 325 articles were included for evaluation. Most studies performed moderately on CLAIM, averaging a score of 28·9±7·5 out of 53, but poorly on FUTURE-AI, averaging 5·1±2·1 out of 30.

Interpretations:

Imaging-AI tools for STBT remain at the proof-of-concept stage, indicating significant room for improvement. Future efforts by AI developers should focus on design (e.g. define unmet clinical need, intended clinical setting and how AI would be integrated in clinical workflow), development (e.g. build on previous work, training with data that reflect real-world usage, explainability), evaluation (e.g. ensuring biases are evaluated and addressed, evaluating AI against current best practices), and the awareness of data reproducibility and availability (making documented code and data publicly available). Following these recommendations could improve clinical translation of AI methods.

Funding:

Hanarth Fonds, ICAI Lab, NIHR, EuCanImage.

**Panel 1**

**Research in context**

**Evidence before this study**

Research on the use of AI in diagnosing and predicting the outcomes of soft-tissue and bone tumours (STBT) is becoming more prevalent. However, the clinical adoption of AI methods in this field remains limited, highlighting a significant gap between AI development and its practical implementation in healthcare settings. Previous reviews focused on the accuracy and performance of published STBT tools, however, did not investigate the quality of research. Recent efforts have introduced guidelines with comprehensive criteria specifically designed for structured reporting and responsible development, deployment, and governance of trustworthy AI in healthcare.

**Added value of this study**

This review examines the methodological quality of published literature by assessing it against two best-practice guidelines, which were chosen to complement each other and cover a wide range of criteria. Aspects related to study quality, study design, and trustworthy and deployable AI, as assessed in this review using the CLAIM and FUTURE-AI guidelines, may be even more important factors than their performance for assessing their potential translation to the clinic. This review highlights what the field is doing well and where future research should focus. The review includes all research using AI methods investigating STBT, giving it a far wider scope than previous reviews. Furthermore, this is a fast-moving field, hence updates on previous reviews are required.

**Implications of all the available evidence**

Currently published AI methods are producing promising proof-of-concept results but are not ready for clinical application. This work highlights opportunities and provides recommendations for AI developers and clinical professionals for future research to drive clinical implementation.

## Introduction

Primary soft-tissue and bone tumours (STBT) are among the rarest neoplasms in humans, comprising both benign and malignant lesions. Malignant STBT, i.e. sarcoma, account for approximately 1% of all neoplasms.[1] These tumours may occur at any age and almost any anatomical site, arising from cells of the connective tissue, including muscles, fat, blood vessels, cartilage, and bones.[2] The rarity of STBT, along with their diverse subtypes and varied clinical behaviour, poses substantial challenges in accurate diagnosis and prognosis.

Radiological imaging (including nuclear medicine) is crucial in evaluating and monitoring STBT. Technological advancements in imaging modalities have led to a substantial increase data volume, along with a corresponding growth in the expertise required for its interpretation. The growing utilisation of radiological imaging and complexity of analysis has increased radiologists' workload. Therefore, developing intelligent computer-aided systems and algorithms for automated image analysis that can achieve faster and more accurate results is crucial.[3] For STBT, intelligent systems may help non-specialised radiologists in diagnosing rare cancers more effectively. Furthermore, an increased caseload is associated with higher interpretive error, which can be avoided with computer-aided diagnostic tools.[4,5]

Artificial intelligence (AI) has become increasingly prevalent in medical image analysis. Over the last 7 years, the number of FDA-approved medical imaging AI products for radiology has substantially increased.[6] However, while medical imaging AI research in STBT has also substantially increased, there are no products developed for STBT among the FDA-approved list.[7] Hence, instead of purely developing novel technological solutions, more research should focus on aligning with areas of unmet clinical need.

Therefore, a systematic assessment of current published research is necessary to identify the issues required to overcome the translational barrier. This systematic review aims to evaluate the existing literature on AI for diagnosis and prognosis of STBT using radiological imaging against two best practice guidelines; CLAIM and FUTURE-AI.[8,9] CLAIM, endorsed by the Radiological Society of North America (RSNA), promotes comprehensive reporting of radiological research that uses AI. FUTURE-AI proposes ethical and technical standards to ensure responsible development, deployment, and governance of trustworthy AI in healthcare. Utilising both guidelines allows for comprehensive coverage of different aspects of AI research.[10] Additionally, this review discusses opportunities for future research to bridge the identified gap between AI research and clinical use in STBT.

## Methods

This systematic review was prospectively registered with PROSPERO (CRD42023467970) and adheres to the Preferred Reporting Items for Systematic Reviews and Meta-analyses (PRISMA) 2020 guidelines.[11] The full study protocol can be found online .[12]

*Search strategy and selection criteria*

Medline, Embase, Web of Science core collection, Google Scholar, and Cochrane Central Register of Controlled Trials were systematically searched for relevant studies. All papers published before 27/09/2023 were included in the initial search; the starting date depended on the coverage of the respective database searched. The detailed search strategy is listed in Appendix 1. The literature search was conducted by the Medical Library, Erasmus MC, Rotterdam, the Netherlands. The database search was repeated on 17/07/2024 to update publications.

Inclusion criteria were: (1) original research papers published in peer-reviewed journals, and (2) studies focusing on radiology-based AI or radiomics characterisation of primary tumours located in bone and/or soft tissues for tasks related to diagnosis or prognosis, e.g. no pure segmentation studies. Exclusion criteria were: (1) animal, cadaveric, or laboratory studies, and (2) not written in English language.

The complete reviewing methodology is illustrated in Figure 1. Three independent reviewers participated in title-and-abstract screening (DS, MM, XW). Retrieved papers were randomly divided into three batches. Reviewers 1 and 2 reviewed one batch, Reviewers 1 and 3 reviewed a second batch, and Reviewers 2 and 3 reviewed the final batch. In cases where there were disagreements in the screening of an abstract, the third reviewer who was not initially involved in reviewing the specific abstract, adjudicated any conflicts.

*Data analysis*

Each paper was scored according to CLAIM and FUTURE-AI guidelines. Checklists were developed based on each guideline. Blank checklists are available in Appendix 2. These guidelines were chosen for their complimentary nature and comprehensive coverage of clinical AI tool requirements.[10]

The CLAIM checklist was adapted from the checklist implemented by Si et. al. to contain more detail in some of the more general checklist items.[8,13,14] CLAIM consists of 44 items, covering the following sections: title, abstract, introduction, methods, results, discussion, and other information. The majority of items focus on the methods (30/44 items). The Methods section is further divided into the following subsections: Study design, Data, Ground truth, Data partition, Testing data, Model, Training, and Evaluation. Similarly, the Results section is divided into Data and Model performance. We further divided three items into twelve sub-items to provide more detailed information. These were: (4) Study objectives and hypotheses (4a and 4b), (7) Data sources (7a-d), and (9) Data preprocessing steps (9a-f). The adapted CLAIM checklist totalled 53 items.

The FUTURE-AI checklist was created from the FUTURE-AI guideline and contains 30 items.[9] These items are split according to the six FUTURE-AI principles: Fairness (3), Universality (4), Traceability (6), Usability (5), Robustness (3), Explainability (2), and General (7). Additionally, FUTURE-AI specifies guidelines for AI tools at various machine learning technology readiness levels (TRL). It recommends (+) or strongly recommends (++) specific guidelines for tools at the proof-of-concept stage (Research) and for those intended for clinical development (Deployable).

All items in both sets of guidelines were scored between 0 and 1, with 0 meaning the item was not addressed, 0·5 meaning it was partially addressed (where relevant and only in FUTURE-AI) and 1 meaning it was fully addressed.

To ensure consistency between scores among reviewers, a subset of papers (n=45) was selected for independent review by all three reviewers. The subset was selected by ordering the papers alphabetically based on the first author's name and choosing the first 45 papers from this order in the initial search. The number of disagreements for each item in either guideline was recorded, and inter-reader variability for each guideline was measured by calculating Fleiss' Kappa statistics ($\kappa$).[15] Fleiss kappa statistics were interpreted according to the guidance given by Fleiss et al., with a score 0–0·4 indicating poor agreement, 0·41–0·75 showing good agreement and >0·75 showing excellent agreement.[15] To construct 95% confidence intervals (95% CI) for the inter-reader variability, 1000× bootstrap resampling was employed. The percentage agreement between all three reviewers was calculated for each item. Following this a consensus discussion was conducted between all three reviewers, allowing discussion and resolution of any systematic differences in interpretation and scoring of specific items. Next, each reviewer re-scored the same subset a second time, several weeks after the first scoring. Kappa statistics and percentage agreements were re-calculated.

After consensus, the remaining included papers were equally divided between the three reviewers and reviewed by a single reviewer. If a reviewer was uncertain how to score a paper, they consulted one or more of the other reviewers for confirmation or discussion. In addition to scoring the CLAIM and FUTURE-AI checklists, the following information was recorded for each paper: (1) year of publication, (2) journal of publication, (3) disease type investigated (soft tissue sarcoma, bone sarcoma, or gastrointestinal stromal tumour – GIST), (4) study design (retrospective or prospective – if a study used both retrospectively and prospectively acquired data it was recorded as being a prospective study), (5) outcome predicted (diagnosis, prognosis, or both), (6) imaging modality (MRI, CT, ultrasound, X-ray, PET-CT, PET-MRI, scintigraphy, or multiple imaging modalities), (7) data source (public, single centre, or multi-centre), and (8) availability of data and AI model source code.

The performance metrics of the corresponding AI models were collected for the top 20 performing papers, as determined by their combined CLAIM and FUTURE-AI scores, that performed external validation. Only the top 20 papers were included for this analysis as reported model performance cannot be reliably reproduced or considered clinically meaningful as low scoring studies lack methodological transparency or do not adhere to best scientific practices. For the same reason, only externally validated papers were selected to ensure robust assessment of model generalizability, reducing the risk of overfitting and dataset-specific bias, thus strengthening the clinical relevance of the reported findings.

*Statistics*

The number of papers adhering to each item of CLAIM/FUTURE-AI was calculated. Descriptive statistics of how well papers scored in each (sub)section/principle were calculated, including mean, standard deviation (SD), maximum, and minimum score, as well as the mean and SD of the guideline adherence rate (AR), which is the score divided by the maximum achievable score.

*Role of Funders*

The funder of the study had no role in study design, data collection, data analysis, data interpretation, or writing of the report.

*Ethics*

This study is a systematic review of published work and thus ethical approval was deemed unnecessary.

## Results

Database searches identified 15,015 published studies, with 5,667 duplicates. After screening, 454 articles were retained for full-text review. After excluding 129 studies a total of 325 unique studies were included in the systematic review (Figure 2). Fifteen of the excluded papers were part of the reproducibility subgroup, meaning 30 articles were independently reviewed by all reviewers. A complete reference list of the final 325 included papers is provided in Appendix 3. Main reasons for exclusion were focusing on different entities (e.g. renal cancer), no use of radiological imaging, or lacking AI-based analysis.

Included studies were published between 2008 and 2024, mostly in the last five years (Figure 3). Of the 325 included studies, most AI methods used hand-crafted imaging features with machine learning (n=221, 68%). Recently, more AI methods used model-learned imaging features (n=62, 19%), i.e. deep learning, or a combination of model-learned and hand-crafted imaging features with machine learning (n=29, 9%). Thirteen studies used hand-crafted imaging features without machine learning.

Study characteristics are illustrated in Figure 4. Disease types included soft tissue tumours (n=125, 38·5%), bone tumours (n=114, 35·1%), and GIST (n=82, 25·2%). Only four studies included both soft tissue and bone tumours (1·2%). Study design was mostly retrospective (n=272, 83.7%), with fewer prospective studies (n=38, 11·7%), and a minority where study design was not clearly documented (n=15, 4·6%). The majority of reports focused on developing AI methods to predict diagnosis (n=206, 63·4%), 109 (33·5%) evaluated prognosis, and 10 (3·1%) studied a combination of diagnosis and prognosis of the disease. Various radiological techniques were evaluated, with 144 (44·3%) studies using MRI, 94 (28·9%) CT, 34 (10·5%) ultrasound, 30 (9·2%) X-ray, 10 (3·1%) PET-CT, 3 (0·9%) PET-MRI, and 1 (0·3%) scintigraphy, and 9 (2·8%) multiple modalities. One-hundred-and-ninety (58·5%) studies collected data from a single centre, whereas 93 (28·6%) utilised imaging from multiple centres. Nineteen studies did not clearly document data provenance (5·8%). Furthermore, 23 (7·1%) studies used publicly available data from two sources (Table 1). AI methods were most often validated with separate internal test data (n=214, 65·8%), and sometimes additionally with external test data (n=70, 21·5%). Several AI methods were not validated with independent data or validation was not clearly documented (n=41, 12·6%). Only 5 (1·5%) studies made data available, with 238 (73·2%) studies not providing or not specifying data availability, and 82 (25·2%) studies stating data would be made available on reasonable request. Similarly, AI source code to facilitate reproducibility was only made available in 23 (7·1%) studies, with 287 (88·3%) not providing or not specifying code availability, and 15 (4·6%) studies indicating code would be made available on reasonable request.

Kappa statistics for inter-reader variability increased from 0·58 (95% CI: [0·55, 0·62]) and 0·68 (95% CI: [0·61, 0·75]) for CLAIM and FUTURE-AI before consensus discussion, to 0·80 (95% CI: [0·78, 0·83]) and 0·92 (95% CI: [0·88, 0·95]) after, showing excellent agreement (Supplementary Figure S1 and S2).

Individual scores for each item in Figure 5 for CLAIM and 6 for FUTURE-AI. Section level scores are provided in Table 2 and 3. Scores by year are available in Supplementary Figure S3 and S4, both showing an increasing trend. Scores by tumour type, method type, and outcome are available in Supplementary Figures S5 and S6, all showing no clear distinction between groups. Individual paper scores for each item are documented in Supplementary Figures S7 and S8, and are also available online as interactive figures and tables.[16]

The included studies performed moderately on the CLAIM checklist, with a mean score of 28·9 out of 53 (SD: 7·5, min–max: 4·0–48·0, AR mean±SD: 55%±14%). All items were reported at least once, but several were only reported in less than 15% of the papers (n≤50 papers) including: define a study hypothesis at the design phase (CLAIM-4b, 13·8%), data de-identification methods (CLAIM-11, 3·4%), how missing data were handled (CLAIM-12, 8·2%), intended sample size and how it was determined (CLAIM-21, 4%), robustness or sensitivity analysis (CLAIM-30, 13·8%), methods for explainability or interpretability (CLAIM-31, 12·9%), registration number and name of registry (CLAIM-34, 2·8%), and documented where full study protocol can be accessed (CLAIM-42, 12·3%).

The included studies rarely adhered to FUTURE-AI, with a mean score of 5·1 out of 30 (SD: 2·1, min–max: 0–11·5, AR: 17%±7%). From the 30 items, 5 were never reported. Only 6 items were partially reported in over half of the reviewed papers (n>162) including: collecting and reporting on individuals' attributes (Fairness-2, 83·1%), using community-defined standards (Universality-2, 56%), defining use and user requirements (Usability-1, 85·2%), engaging interdisciplinary stakeholders (General-1, 86·2%), implementing measures for data privacy and security (General-2, 85·2%), and defining an adequate evaluation plan (General-4, 67·7%).

Strongly recommended items by FUTURE-AI for proof-of-concept AI studies (Research), were reported more frequently than recommended items, with mean scores of 2·9 out of 12 (SD: 1·1, min–max: 0–7, AR: 24%±9%) and 2·3 out of 16 (SD: 1·2, min–max: 0–6·5, AR: 14%±8%), respectively. However, this trend was not observed in items intended to assess studies for clinical deployability (Deployable), where the mean scores were 3·8 out of 24 (SD: 1·7, min–max: 0–10, AR: 16%±7%) for strongly recommended items and 1·3 out of 4 (SD: 0·7, min–max: 0–3, AR: 33%±18%) for recommended items.

Performance measurements of the top 20 performing papers (summed score of both CLAIM and FUTURE-AI) which included external validation are provided in Table 4. These studies covered diverse disease types (soft-tissue tumours: n=9, bone tumours: n=8, GIST: n=3), imaging modalities (MRI: n=11, CT: n=4, X-ray: n=4, ultrasound: n=1), outcomes (diagnosis: n = 12, prognosis: n= 7 and both diagnosis and prognosis: n =1), and AI methodologies (machine learning model using a combination of hand-crafted and model-learned imaging features: n=3; machine learning using model-learned features: n=6; machine learning using hand-crafted imaging features: n=11). Overall, AI methods demonstrated strong performance for their respective tasks, however there is a wide range in performance between models (AUC range: 0·64–0·95). However, most studies relied on a single centre for external validation (n=12), and only a few included prospective validation (n=2). These studies had a mean score of 40·4 out of 53 (SD: 3·0, AR mean±SD: 76%±5·8%) for CLAIM and 8·4 out of 30 (SD: 1·6, AR mean±SD: 28%±5·4%) for FUTURE-AI. Finally, among these top 20 studies, we explored potential associations between performance metrics, individual guideline scores, and three main study categories, as summarized in Supplementary Table S1. This showed no obvious differences in scores and performance metrics between any of the groups.

**Discussion**

This work has systematically identified and summarised radiological imaging-AI research on STBT and conducted comprehensive evaluation of published literature against two best-practice guidelines: CLAIM and FUTURE-AI. These guidelines were developed to ensure that AI tools target unmet clinical needs, are transferrable, generalisable, and can be used in real-world clinical practice. Analysis revealed a rapid increase in experimental AI tools for imaging-based STBT evaluation over the past five years. Studies performed moderately against CLAIM (28·9±7·5 out of 53) and poorly against FUTURE-AI evaluations (5·1±2·1 out of 30). The poor results in FUTURE-AI are expected as these guidelines are recent and set high requirements. Several papers do show higher scores in both CLAIM and FUTURE-AI (Table 4) and show promising results in external validation cohorts (AUC range: 0·784-0·948). However, the highest scoring paper achieved only a 11·5 out of 30 in FUTURE-AI, highlighting room for improvement. These results suggest that while progress has been made in developing AI tools for STBT, most studies are still at the proof-of-concept stage and there remains substantial room for improvement to guide future clinical translation. Panel 2 summarises the authors' recommendations, focusing on five keys topics: design, development, evaluation, reproducibility, and data availability.

In the design stage, several critical aspects warrant more attention. Intended clinical settings (Universality-1) and prior hypotheses (CLAIM-4b) should be reported. On a positive note, over 85% of studies involved interdisciplinary teams (Usability-1, General-1), which is recommended for effective AI tool development.[9] However, most studies did not comprehensively identify possible sources of bias at an early stage (Fairness-1, Robustness-1), which could limit the applicability of these AI tools. To overcome this, interdisciplinary stakeholders should work together from the design stage to identify the clinical role of the AI tool, ensure it integrates into the clinical workflow, and any possible sources of bias.

In the development stage, studies generally reported dataset source and conducted research with appropriate ethical approvals (CLAIM-7). However, almost half of studies did not assess biases during AI development (Fairness-3) and very few studies trained with representative real-world data (Robustness-2), which can hinder the transferability of AI tools, especially given the highly heterogeneous imaging characteristics of STBT. Another notable gap is a lack of focus on explainability and traceability. Few studies addressed items under FUTURE-AI Explainability (1-2) and Traceability (1-3), similar shortcoming was observed in the CLAIM checklist (CLAIM-31). While accuracy is crucial in medical practice, it is often argued that AI methods should go beyond pure performance metrics by addressing other factors such as prediction uncertainties, explaining their outputs, and providing clinicians with detailed information.[17] For AI tools to be effective in clinical decision-making, explainability is vital to ensure clinicians understand and can trust the AI's reasoning.[18] Additionally, to assist with AI development, research should build on previous work where possible. To assist with this, researchers should continue to adhere to community-defined standards, which is currently done in over half of the reviewed papers, and ensure their code is available. This review shows that almost all included studies developed new models rather than adapting or enhancing existing ones, even when promising results were achieved. Finally, it is integral that AI tools are easy for the end-user to use in the clinical workflow, however only two studies developed a graphical user interface for user experience testing (Usability-3).[19,20]

Regarding evaluation, while over 85% of studies adopted relevant metrics and reported AI algorithm performance (CLAIM-28 and 37), only 22% conducted external validation (CLAIM-33), and most used single-institute datasets (Universality-3). Furthermore, several studies lacked thorough internal validation (Robustness-3, General-4). AI tools should be tested against independent external data, ideally from multiple sources, to assess the tool's universality and prevent site-specific bias. Accuracy metrics should also be compared against current best-practice (i.e. compared to radiologists) to ensure AI tools offer improvements in outcomes. Less than 20% of studies reported failure analysis or incorrectly classified cases (CLAIM-39). Including failure analysis is crucial to identify potential pitfalls, helping users understand when it is appropriate to use the tool. Developers should also ensure that the tool is robust against the biases identified during the design stage.

Regarding reproducibility, most studies fail to provide adequate materials (code, model, and data) to reproduce published results. Only around 10% of studies offered a full study protocol, including comprehensive methodology or code. Making protocols and code available enables others to reproduce the study across multiple steps, such as data preprocessing, ground truth acquisition, model construction, and training procedure. The lack of accessible and reproducible AI research in STBT could impede the adoption of these tools, as sarcoma centres may struggle to reproduce the tools performance locally. Adhering to guidelines such as CLAIM could enhance the quality and accessibility of these protocols.

Regarding data availability, there is a lack of freely accessible annotated imaging datasets of STBT, as highlighted in Table 1. Although 25% of published research stated that data used was available by request, a recent study by Gabelica et al. (2022) investigating compliance with data sharing statements showed a response rate of 14%, with only 6·8% supplying the data.[21] One challenge in creating these datasets is the time required and the need for an easy-to-use format. Structured and standardised reporting in clinical practice could help reduce the effort needed for retrospective data collection. However, AI developers often struggle to collate data themselves, especially since STBT are rare and only treated at tertiary sarcoma centres. This underscores the importance of collaborating with clinical professionals. Increasing data availability would accelerate AI tool development and allow for external validation of models. Potential solutions include hosting "grand challenges" where clinicians provide data for AI developers to tackle a real-world clinical problem, or employing federated learning, which has proven effective for training AI models on rare tumours across international networks.[22-24]

Several reviews described the use of AI or radiomics in STBT management.[25-28] This study expands and complements these previous reviews, including a substantially larger volume of included publications (325 vs. 21-52 reports) primarily due to our extended scope and search strategy, including benign soft-tissue tumours, bone tumours, and a broad range of AI methods (i.e. not limiting to radiomics with hand-crafted features). Furthermore, most previous reviews only examined the accuracy and performance of published AI tools in the field; the current systematic review instead examined the methodological quality of published literature by assessing this against best-practice guidelines. The only other systematic reviews that, to the authors knowledge, have assessed quality of AI research in radiology imaging for STBT are Crombé et al. (2020) (52 studies) and De Angelis et al. (2024) (49 studies), both scoring against the Radiomics Quality Score (RQS).[25,26] In this study, different scoring systems were deliberately chosen as CLAIM and FUTURE-AI are independent but complementary guidelines, providing a broader assessment of overall quality than using only one.[10] FUTURE-AI allows assessment of trustworthiness, deployability, and translation to clinical practice, while CLAIM guidelines, which are endorsed by the RSNA, ensures that studies are reported according to a standard set of information especially designed for medical imaging AI. Findings indicate that the field continues to produce promising proof-of-concept results but is not ready to make the jump to clinical application. This agrees with earlier work in the field.

To better understand the relationship between adherence to reporting guidelines and model performance, we examined the top 20 studies with the highest combined CLAIM and FUTURE-AI scores. Our analyses suggest that no particular subfield demonstrates consistently superior performance, with reported metrics varying widely—even among similar models. This underscores the need for further external validation and standardization. Whilst some studies show promising results, the overall heterogeneity highlights the complexity of AI performance assessment.

Subgroup analysis in which CLAIM and FUTURE-AI scores were investigated by tumour type, method type and outcome, showed no obvious differences between groups although papers performing statistics on hand crafted features scored worse than studies which used some form of machine learning. This is not surprising as the guidelines we chose focus on the use of AI. There was a general trend for a small increase in scores for both guidelines over time. This implies that whilst the quality of AI-based research is improving over time no field assessed in this review is ahead than any other.

There are limitations to this study. First, due to the large volume of literature, most papers were scored by a single reviewer. However, a sub-group of papers were scored by three reviewers followed by consensus analysis, showing excellent agreement, and reviewers remained in discussion if they had doubts about how best to score a paper for a particular category. Two or more reviewers per paper might have provided more robust results but would have required a significant time investment for likely only marginal gains. Secondly, in the reproducibility study with subgroups, papers were selected by alphabetical order based on the first author's name. While this approach introduces a degree of randomness, a fully randomised selection process would have been more robust to minimise potential biases. Third, there are other scoring guidelines such as APPRAISE AI, TRIPOD-AI, or RQS.[51-53] Future studies could benefit from integrating other frameworks, other than CLAIM and FUTURE-AI, to provide a more comprehensive evaluation of both reporting adherence and study quality

In conclusion, this review discusses the growing volume of published work evaluating imaging-related AI tools to aid in diagnosis, prognosis, and management of soft tissue and bone tumours. The top performing papers, as determined by both guidelines, may represent encouraging steps toward bringing AI in radiology closer to clinical translation, however even these have some limitations. The identified limitations of the reviewed studies with respect to CLAIM and FUTURE-AI guidelines will need to be addressed before such tools can translate into the clinical domain. Several opportunities have been identified and the authors' recommendations to promote translation of AI methods into clinical practice are summarised in Panel 2. Addressing these points may help drive clinical adoption of AI tools into the radiology workflow in a responsible and effective way.

**Contributors**

D.J.S., M.M., X.W.: conceptualisation, data curation, formal analysis, investigation, methodology, project administration, visualisation, writing – original draft, and writing – review & editing; S.K., M.P.A.S: conceptualisation, investigation, methodology, supervision, writing – review & editing; A.F.S., P.R., E.H.G.O., J.J.V., R.H., K.L., D.F.H., G.J.L.H.L., C.V., D.J.G., W.J.N.: methodology, supervision, writing – review & editing; S.K., M.P.A.S., M.M., E.H.G.O., J.J.V., C.V., D.J.G., W.J.N.: funding acquisition. All authors read and approved the final version of the manuscript. D.J.S, M.M., X.W., S.K., M.P.A.S. have accessed and verified the data. D.J.S, M.M., X.W. have contributed equally. S.K. and M.P.A.S. have contributed equally.

**Data Sharing Statement**

Empty checklists for this review are included in the supplementary material. All data collected and analysed in this study are available online.[16] A website (https://douwe-spaanderman.github.io/AI-STTandBoneTumour-Review/) with interactive figures and tables with scores for each paper is also available online.

**Declaration of Interests**

WJN is the founder of Quantib and was scientific lead until 31-1-2023. JJV received a grant to institution from Qure.ai / Enlitic; consulting fees from Tegus; payment to institution for lectures from Roche; travel grant from Qure.ai; participation on a data safety monitoring board or advisory board from Contextflow, Noaber Foundation, and NLC Ventures; leadership or fiduciary role on the steering committee of the PINPOINT Project (payment to institution from AstraZeneca) and RSNA Common Data Elements Steering Committee (unpaid); phantom shares in Contextflow and Quibim; chair scientific committee EuSoMII (unpaid); chair ESR value-based radiology subcommittee (unpaid); member editorial board European Journal of Radiology (unpaid). SK and EHGO are scientific directors of the ICAI lab "Trustworthy AI for MRI", a public-private research program partially funded by General Electric Healthcare. The other authors do not have any conflicts of interest.

**Acknowledgments**

This research was supported by an unrestricted grant of Stichting Hanarth Fonds, The Netherlands. MPAS and SK acknowledge funding from the research project EuCanImage (European Union's Horizon 2020 research and innovation programme under grant agreement Nr. 95210). MPAS also acknowledges funding from a NGF AiNed Fellowship (NGF.1607.22.025). MM, Doctoral Clinical and Practitioner Academic Fellow, NIHR302901, is funded by Health Education England (HEE) / National Institute for Health and Research (NIHR) for this research project. This research was conducted within the "Trustworthy AI for MRI" ICAI lab within the project ROBUST, funded by the Dutch Research Council (NWO), GE Healthcare, and the Dutch Ministry of Economic Affairs and Climate Policy (EZK). AS receives salary support from ICNIHR Leeds Biomedical Research Centre (NIHR203331). The funders had no role in study design, data collection and analysis, decision to publish, or preparation of the manuscript. The views expressed in this publication are those of the author and not necessarily those of the NIHR, NHS or the UK Department of Health and Social Care.

**Panel 2: Recommendations to promote clinical translation of AI methods for soft-tissue and bone tumours.**

**Design**
- Interdisciplinary stakeholders should define; (A) the unmet clinical need, (B) the intended use of AI, (C) intended clinical setting in which AI should operate, (D) the end-user requirements, (E) how AI would operate in clinical workflow.
- Possible types and sources of bias (e.g. sex, age, ethnicity, socioeconomics, geography) should be identified at the early design stage.

**Development**
- Data used for AI development should reflect real-world data used in the intended clinical setting or preferably retrieved from the clinical setting. Additionally, sources of variation and potential biases should be investigated early in the development process.
- Explainability of AI methods should be developed and implemented in a way that it is possible to understand why an AI tool has arrived at its predictions.
- AI development should build on previous work by: (A) adhering to community-defined standards, and (B) considering previous existing methods by validating or improving them whenever possible.
- Ensure that AI tools are easy for the end-user to use in a clinical setting.

**Evaluation**
- AI tools should be evaluated using independent external test data. Limits on universality of the external test sets should be discussed.
- AI tools should be evaluated against current best practices, e.g. classification by radiologist or histology results from biopsy, and evaluated with intended end-users.
- Failure analysis of incorrect classified cases should be conducted.
- The robustness and sensitivity to variations and biases in data, identified prior to AI development, should be thoroughly investigated.

**Reproducibility**
- Code should be made publicly available, readable, usable and traceable to increase confidence in the method.
- The Methods section should comprehensively cover all aspects of AI development, including; (A) data preprocessing, (B) ground truth acquisition, (C) a detailed description of the AI methodology, and (D) the training procedures. To this end, the Checklist for Artificial Intelligence in Medical Imaging (CLAIM) could be followed.

**Data availability**
- Structured and standardised reporting should be introduced in clinical practice to limit the manual work required in retrospective data collection.
- Tertiary sarcoma centres should collect labelled data and make this publicly available, preferably in the context of a "grand challenge", while protecting patient details and respecting privacy.
- To protect patient privacy and avoid excessive data-sharing, researchers could work together using a federated learning approach.

**Tables**

**Table 1**: Open-access datasets available with imaging for soft-tissue and bone tumours.

| Data | **Vallières el al. (2015) [29]** | **Starmans et al. (2021) [preprint - 30]** |
|---|---|---|
| **Origin** | Canada | the Netherlands |
| **Disease type** | Various soft-tissue sarcoma (Extremities) | Various soft-tissue tumours |
| **Imaging modality** | MR and PET-CT | MR or CT |
| **Number of patients** | 51 | 564 |
| **Additional data** | Tumour segmentation and clinical outcome (lung metastasis) | Tumour segmentation and clinical outcome (phenotype) |

**Table 2**: Summary scores of the included studies for each (sub)section of the Checklist for Artificial Intelligence in Medical Imaging (CLAIM).

| (Sub)section | Maximum achievable score | Score (Mean ± SD) | Max score | Min score | Adherence rate (Mean ± SD) |
| --- | --- | --- | --- | --- | --- |
| **Title / Abstract** | **2·0** | **2·0 ± 0·2** | **2·0** | **0·0** | **98% ± 12%** |
| **Introduction** | **3·0** | **2·1 ± 0·4** | **3·0** | **0·0** | **70% ± 14%** |
| **Methods** | **38·0** | **19·8 ± 5·8** | **34·0** | **0·0** | **52% ± 15%** |
| Study design | 2·0 | 1·8 ± 0·5 | 2·0 | 0·0 | 89% ± 24% |
| Data | 15·0 | 8·0 ± 2·8 | 14·0 | 0·0 | 54% ± 18% |
| Ground truth | 5·0 | 2·9 ± 1·4 | 5·0 | 0·0 | 57% ± 29% |
| Data partitions | 2·0 | 1·7 ± 0·6 | 2·0 | 0·0 | 87% ± 30% |
| Testing data | 1·0 | 0·0 ± 0·2 | 1·0 | 0·0 | 4% ± 20% |
| Model | 3·0 | 1·5 ± 1·0 | 3·0 | 0·0 | 51% ± 33% |
| Training | 3·0 | 1·2 ± 0·9 | 3·0 | 0·0 | 40% ± 31% |
| Evaluation | 7·0 | 2·7 ± 1·3 | 6·0 | 0·0 | 38% ± 18% |
| **Results** | **5·0** | **2·6 ± 1·2** | **5·0** | **0·0** | **52% ± 24%** |
| Data | 2·0 | 1·0 ± 0·8 | 2·0 | 0·0 | 50% ± 39% |
| Model performance | 3·0 | 1·6 ± 0·8 | 3·0 | 0·0 | 53% ± 25% |
| **Discussion** | **2·0** | **1·3 ± 0·6** | **2·0** | **0·0** | **66% ± 32%** |
| **Other information** | **3·0** | **1·2 ± 0·9** | **3·0** | **0·0** | **39% ± 31%** |
| **Overall** | **53·0** | **28·9 ± 7·5** | **48·0** | **4·0** | **55% ± 14%** |

**Table 3**: Summary scores of the included studies for each principle from the FUTURE-AI international consensus guideline for trustworthy and deployable AI.

| Principle | Maximum achievable score | Score (Mean ± SD) | Max Score | Min Score | Adherence rate (Mean ± SD) |
|---|---|---|---|---|---|
| **Fairness** | 3·0 | 1·1 ± 0·7 | 2·5 | 0·0 | 37% ± 22% |
| **Universality** | 4·0 | 0·8 ± 0·7 | 3·0 | 0·0 | 20% ± 17% |
| **Traceability** | 6·0 | 0·1 ± 0·2 | 1·0 | 0·0 | 1% ± 3% |
| **Usability** | 5·0 | 0·5 ± 0·3 | 3·0 | 0·0 | 10% ± 7% |
| **Robustness** | 3·0 | 0·4 ± 0·4 | 2·5 | 0·0 | 14% ± 12% |
| **Explainability** | 2·0 | 0·1 ± 0·2 | 1·5 | 0·0 | 4% ± 12% |
| **General** | 7·0 | 2·2 ± 0·8 | 3·5 | 0·0 | 32% ± 11% |
| **Overall** | 30·0 | 5·1 ± 2·1 | 11·5 | 0·0 | 17% ± 7% |

**Table 4:** Performance measurements of the top 20 performing papers, as determined by their combined CLAIM and FUTURE-AI scores, among those that performed external validation. AUC = area under the curve, CI = confidence interval, NPV = negative predictive value. *AI development centre was also included as one of the eight external validation centres.
† Values are mean ± standard deviation

| Author | Short description | Validation | Performance (Proportion, 95% CI) |
|---|---|---|---|
| Ye et al. [31] | A multi-task machine learning model using learned imaging features (deep learning) for the segmentation, detection, and differentiation of malignant and benign primary bone tumours, as well as bone infections, leveraging multi-modal inputs including T1-weighted MRI, T2-weighted MRI, and clinical data. | *External validation* 53 patients from 1 centre | AUC: 0·900 (0·773–1·000) Accuracy: 0·783 (0·581–0·903) Sensitivity: 0·756 (0·552–0·886) Specificity: 0·886 (0·764–0·950) |
| Dong et al. [32] | Machine learning model using learned imaging features (deep learning) differentiating gastrointestinal stromal tumours (GISTs) and leiomyomas on endoscopic ultrasonography. | *External validation* 241 patients from 1 centre  *Prospective validation* 59 patients from 1 centre | *External validation* AUC: 0·948 (0·921–0·969) Accuracy: 0·917 (0·875–0·946) Sensitivity: 0·903 (0·834–0·945) Specificity: 0·930 (0·872–0·963) Precision: 0·919 (0·853–0·957) NPV: 0·915 (0·855–0·952)  *Prospective validation (for GISTs and leiomyomas, respectively)* AUC: 0·865 (0·782–0·977) and 0·864 (0·762–0·966) Accuracy: 0·865 and 0·864 Sensitivity: 0·897 and 0·857 Specificity: 0·833 and 0·871 Precision: 0·839 and 0·857 NPV: 0·893 and 0·881 |
| Xie et al. [33] | Machine learning model using learned imaging features (deep learning) to classify histological types of primary bone tumours on radiographs. | *External validation* 89 patients from 1 centre | AUC: 0·873 (0·812–0·920) Accuracy: 0·687 (0·614–0·783) Sensitivity: 0·572 (0·457–0·685) Specificity: 0·916 (0·893–0·938) |
| Xu et al. [34] | Machine learning model using a combination of hand-crafted and model-learned imaging features to differentiate between retroperitoneal lipomas and well-differentiated liposarcomas based on MDM2 status on contrast-enhanced CT. | *External validation* 63 patients from 2 centre | AUC: 0·861 (0·737–0·985) Accuracy: 0·810 |
| Arthur et al. [35] | Machine learning model using hand-crafted imaging features classifying histological type and tumour grade in retroperitoneal sarcoma on CT. | *External validation* 89 patients from 8 centres* | *Histology and Grade* AUC: 0·928 and 0·882 Accuracy: 0·843 and 0·823 Sensitivity: 0·923 and 0·800 Specificity: 0·829 and 0·848 Precision: 0·480 and 0·865 NPV: 0·984, 0·778 |
| Guo et al. [36] | Machine learning model using a combination of hand-crafted and model-learned imaging features to classify histological grade and predict prognosis of soft-tissue tumours on MRI. | *External validation* 125 and 44 patients from 2 centres  *Prospective validation* 12 patients from 1 centre | *External validation (Centre 1 and Centre 2)* AUC: 0·860 (0·787–0·916) and 0·838 (0·696–0·932) Accuracy: 0·840 and 0·750 Sensitivity: 0·835 and 0·840 Specificity: 0·794 and 0·737 Hazard ratio: 4·624 (1·924–11·110) and 2·920 (0·603–14·150)  *Prospective validation* AUC: 0·819 (0·501–0·974) Accuracy: 0·667 Sensitivity: 0·667 Specificity: 1·000 |

| Study | Description | Validation | Results |
|---|---|---|---|
| Gitto et al. [37] | Machine learning model using hand-crafted imaging features differentiating atypical cartilaginous tumour and grade II chondrosarcoma of long bones on MRI. | *External validation* 65 patients from 1 centre | AUC: 0·94 for atypical cartilaginous tumour and 0·90 for grade II chondrosarcomal<br>Accuracy: 0·92<br>Sensitivity: 0·92<br>Precision: 0·92 |
| Von Schaky et al. [38] | Machine learning model using hand-crafted imaging features to distinguish between benign and malignant bone lesions on radiography. | *External validation* 96 patients from 1 centre | AUC: 0·90<br>Accuracy: 0·75 (0·65–0·83)<br>Sensitivity: 0·90 (0·74–0·98)<br>Specificity: 0·68 (0·55–0·79)<br>Precision: 0·57 (0·42–0·71)<br>NPV: 0·94 (0·82–0·99) |
| Gitto et al. [39] | Machine learning model using hand-crafted imaging features differentiating atypical cartilaginous tumour and high-grade chondrosarcoma of long bones on radiography. | *External validation* 30 patients from 1 centre | AUC: 0·90<br>Accuracy: 0·80<br>Sensitivity: 0·89<br>Specificity: 0·67 |
| Cao et al. [40] | Machine learning model using hand-crafted imaging features predicting the local recurrence after surgical treatment of primary dermatofibrosarcoma protuberans, based on MRI. | *External validation* 42 patients from 1 centre | AUC: 0·865 (0·732–0·998) for 3-year and 0·931 (0·849–1·00) for 5 year<br>C-index: 0·866 (0·786–0·946) |
| Yang et al. [41] | Machine learning model using hand-crafted imaging features predicting progression-free survival after imatinib therapy in patients with liver metastatic gastrointestinal stromal tumours on multi-sequence MRI. | *External validation* 45 patients from 1 centre | AUC: 0·766 for 1-year, 0·776 for 3-year, and 0·893 for 5-year<br>C-index: 0·718 (0·618–0·818) |
| Chen et al. [42] | Machine learning model using hand-crafted imaging features predicting pathologic response to neoadjuvant chemotherapy (NAC) in patients with osteosarcoma on MRI. | *External validation* 34 patients from 3 centres | AUC: 0·842 (0·793–0·883)<br>Accuracy: 0·765 ± 0·020†<br>Sensitivity: 0·739 ± 0·032†<br>Specificity: 0·909 ± 0·026† |
| Liang et al. [43] | Machine learning model using a combination of hand-crafted and model-learned imaging features for predicting lung metastases in patients with soft-tissue sarcoma on MRI. | *External validation* 126 patients from 2 centre | AUC: 0·833 (0·732–0·933)<br>Accuracy: 0·897<br>Sensitivity: 0·474<br>Specificity: 0·972<br>Precision: 0·750<br>NPV: 0·912 |
| Kang et al. [44] | Machine learning model using learned imaging features (deep learning) to predict preoperative risk of gastrointestinal stromal tumours on CT. | *External validation* 388 patients from 1 centre | *Low-malignant, intermediate-malignant, and high-malignant*<br>AUC: 0·87 (0·83–0·91), 0·64 (0·60–0·68), and 0·85 (0·81–0·89)<br>Accuracy: 0·81 (0·77–0·85), 0·75 (0·71–0·79), and 0·77 (0·73–0·81)<br>Sensitivity: 0·72 (0·64–0·79), 0·24 (0·14–0·34), and 0·79 (0·73–0·85)<br>Specificity: 0·86 (0·83–0·90), 0·86 (0·82–0·90), and 0·75 (0·70–0·81) |
| He et al. [45] | Machine learning model using learned imaging features (deep learning) for classification of benign, intermediate or malignant primary bone tumours on radiography. | *External validation* 291 patients from 2 centre | AUC: 0·877 (0·833–0·918) benign vs not benign and 0·916 (0·877–0·949) malignant vs not malignant<br>Accuracy: 0·734 |
| Peeken et al. [46] | Machine learning model using hand-crafted imaging features from different timepoints (delta radiomics) predicting pathologic complete response to neoadjuvant therapy in high grade soft tissue sarcoma of trunk and extremity, based on MRI. | *External validation* 53 patients from 1 centre | AUC: 0·75 (0·56–0·93)<br>Accuracy: 0·86<br>Balanced accuracy: 0·57<br>Sensitivity: 0·20<br>Specificity: 0·95<br>Precision: 0·33<br>NPV: 0·90 |
| Foreman et al. [47] | Machine learning model using hand-crafted imaging features predicting the MDM2 gene amplification status in order to differentiate between atypical lipomatous tumours (ALT) and lipomas on MRI. | *External validation* 50 patients from 1 centre | AUC: 0·88 (0·85–0·91)<br>Accuracy: 0·76<br>Sensitivity: 0·70<br>Specificity: 0·81 |

| Spraker et al. [48] | Machine learning model using hand-crafted imaging features predicting overall survival of grade II and III soft-tissue tumours on MRI. | *External validation* 61 patients from 1 centre | Sensitivity: 0·79 Specificity: 0·68 C-index: 0·78 Hazard ratio: 2·4 |
|---|---|---|---|
| Fradet et al. [49] | Machine learning model using a combination of hand-crafted and model-learned imaging features predicting malignancy for lipomatous soft-tissue lesions on MRI. | *External validation* 60 patients from 35 centres | AUC: 0·80 Specificity: 0·63 |
| Gitto et al. [50] | Machine learning model using hand-crafted imaging features differentiating atypical cartilaginous tumours and high-grade chondrosarcomas of long bones on CT. | *External validation* 36 patients from 1 centre | AUC: 0·784 Accuracy: 0·75 |

# Supplementary Table

**Table S1:** Score analysis for different predicted outcomes, disease types, and AI methods of the top 20 studies in terms of highest combined CLAIM and FUTURE-AI score.

| Categories | | N | Mean CLAIM score | Mean FUTURE-AI score | AUC range | Accuracy range | Sensitivity range | Specificity range |
|---|---|---|---|---|---|---|---|---|
| Outcome type | Diagnosis | 12 | 41·2 | 8·7 | 0·78-0·95 | 0·69-0·92 | 0·57-1·00 | 0·63-0·93 |
| | Prognosis | 7 | 39·0 | 7·9 | 0·64-0·93 | 0·77-0·90 | 0·20-0·79 | 0·68-0·97 |
| | Both | 1 | 40·0 | 9·0 | 0·82-0·86 | 0·67-0·84 | 0·67-0·84 | 0·74-1·00 |
| Disease type[a] | Bone tumour | 8 | 41·4 | 8·1 | 0·78-0·94 | 0·69-0·92 | 0·57-0·90 | 0·67-0·92 |
| | Soft-tissue tumour | 9 | 38·9 | 8·6 | 0·75-0·93 | 0·67-0·90 | 0·20-1·00 | 0·63-1·00 |
| | GIST | 3 | 42·0 | 8·7 | 0·64-0·95 | 0·75-0·92 | 0·24-0·90 | 0·75-0·93 |
| Method type | Hand-crafted features | 11 | 39·5 | 7·6 | 0·75-0·94 | 0·75-0·92 | 0·20-0·92 | 0·67-0·95 |
| | Model-learned features | 6 | 42·8 | 9·3 | 0·64-0·95 | 0·67-0·92 | 0·24-0·90 | 0·74-1·00 |
| | Combined hand-crafted and model-learned features | 3 | 38·3 | 9·5 | 0·80-0·86 | 0·80-0·86 | 0·47-1·00 | 0·63-0·97 |

The ranges presented in the table are derived from the minimum and maximum values reported for each metric across the selected studies. For the study categorized under 'both,' performance metrics were reported from three external validation sites, contributing to the observed ranges.

[a] No papers investigating both Soft-tissue tumour (STT) and Bone tumours were in the top 20 scoring papers.

**Figures**

**Figure 1:** Reviewing methodology.

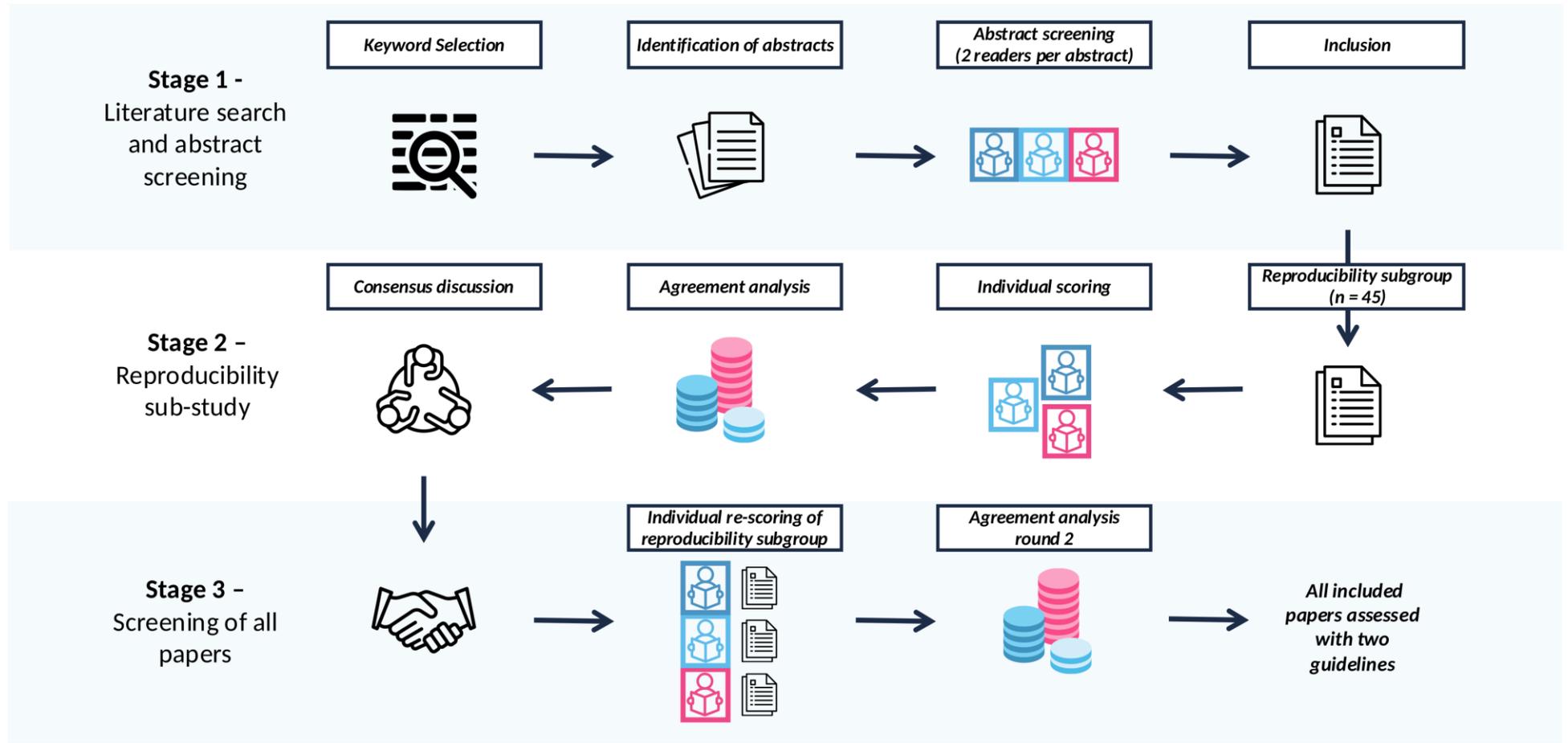

**Figure 2:** PRISMA flow diagram.
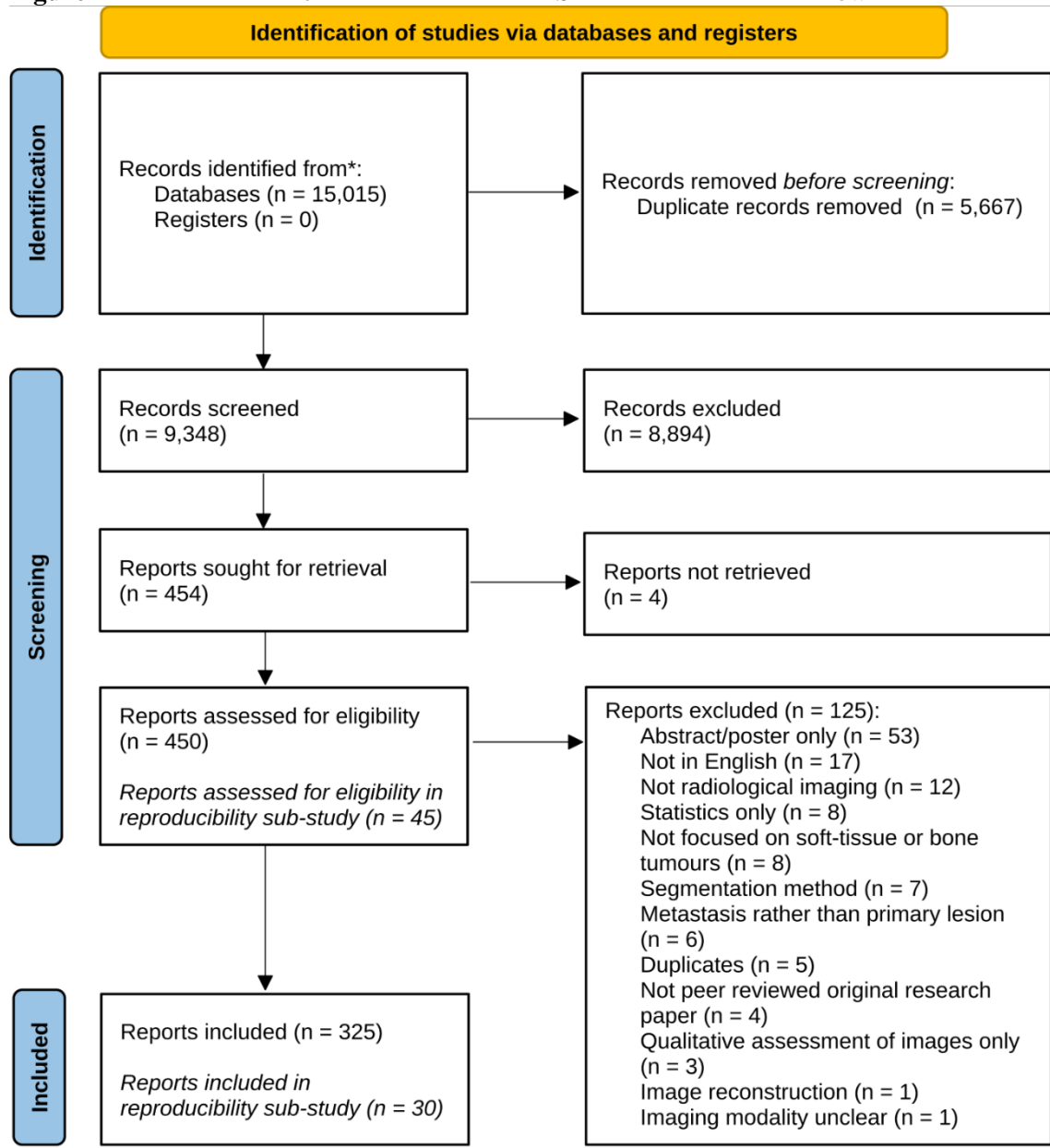

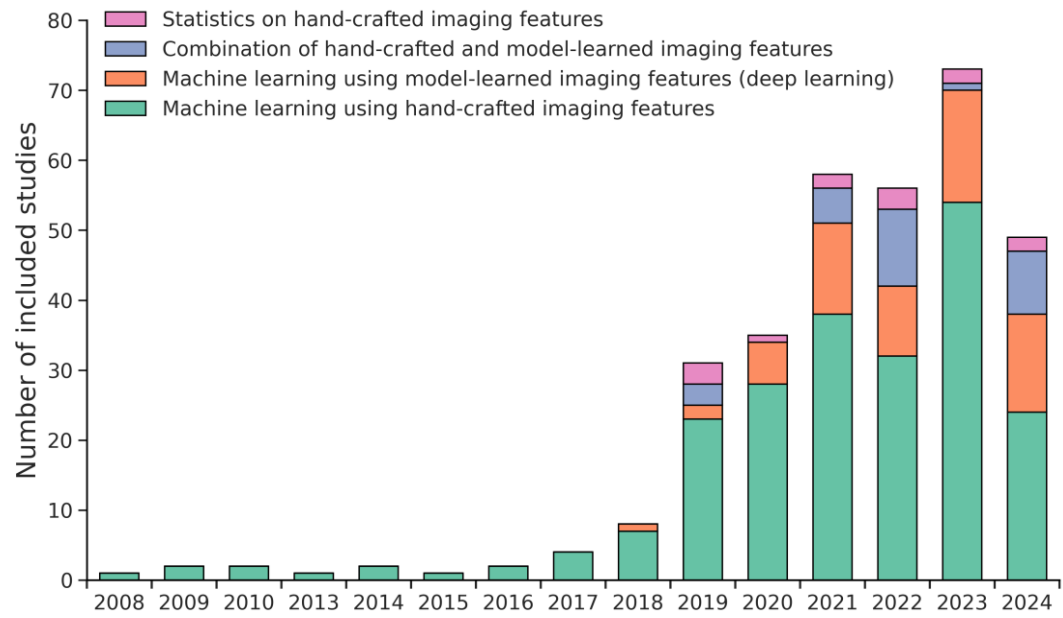

**Figure 3:** Number of included studies (n=325) between 2008 and July 2024, color coded for the various AI methodologies used.

**Figure 4:** Characteristics of the studies included (n=325) as percentages.

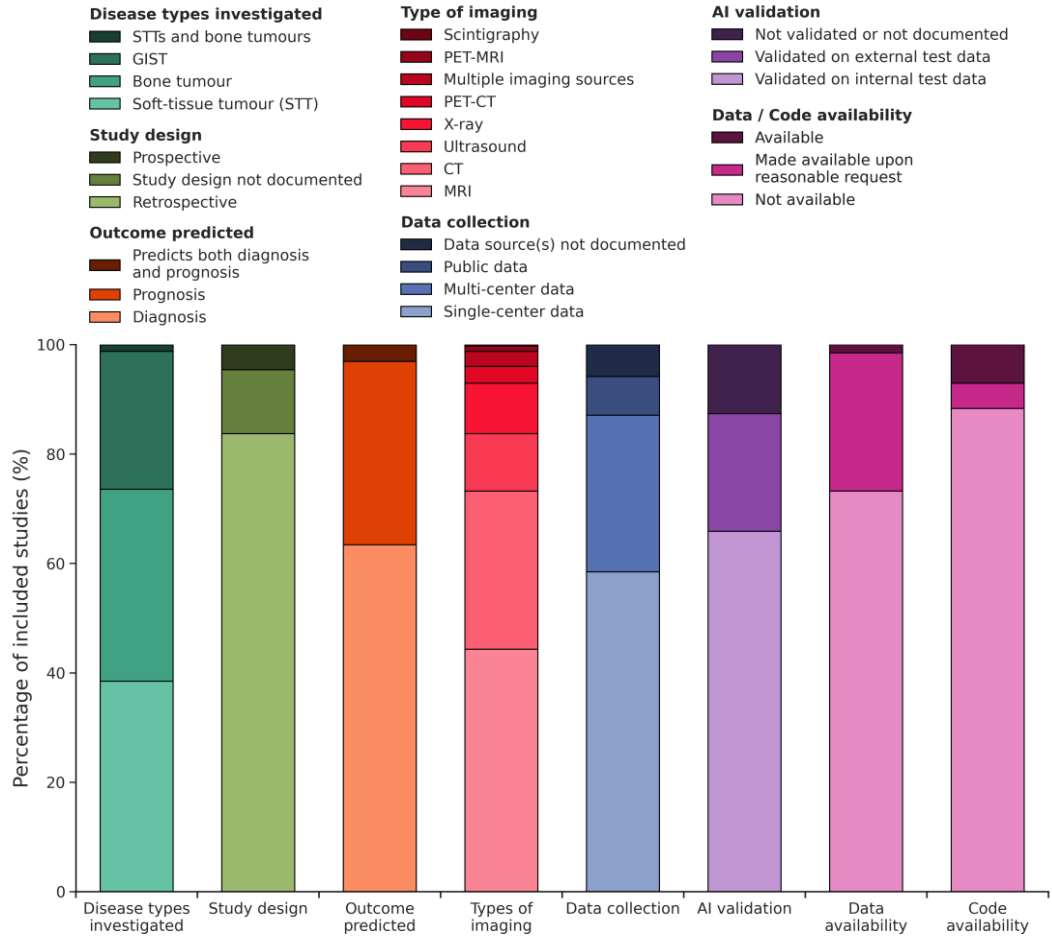

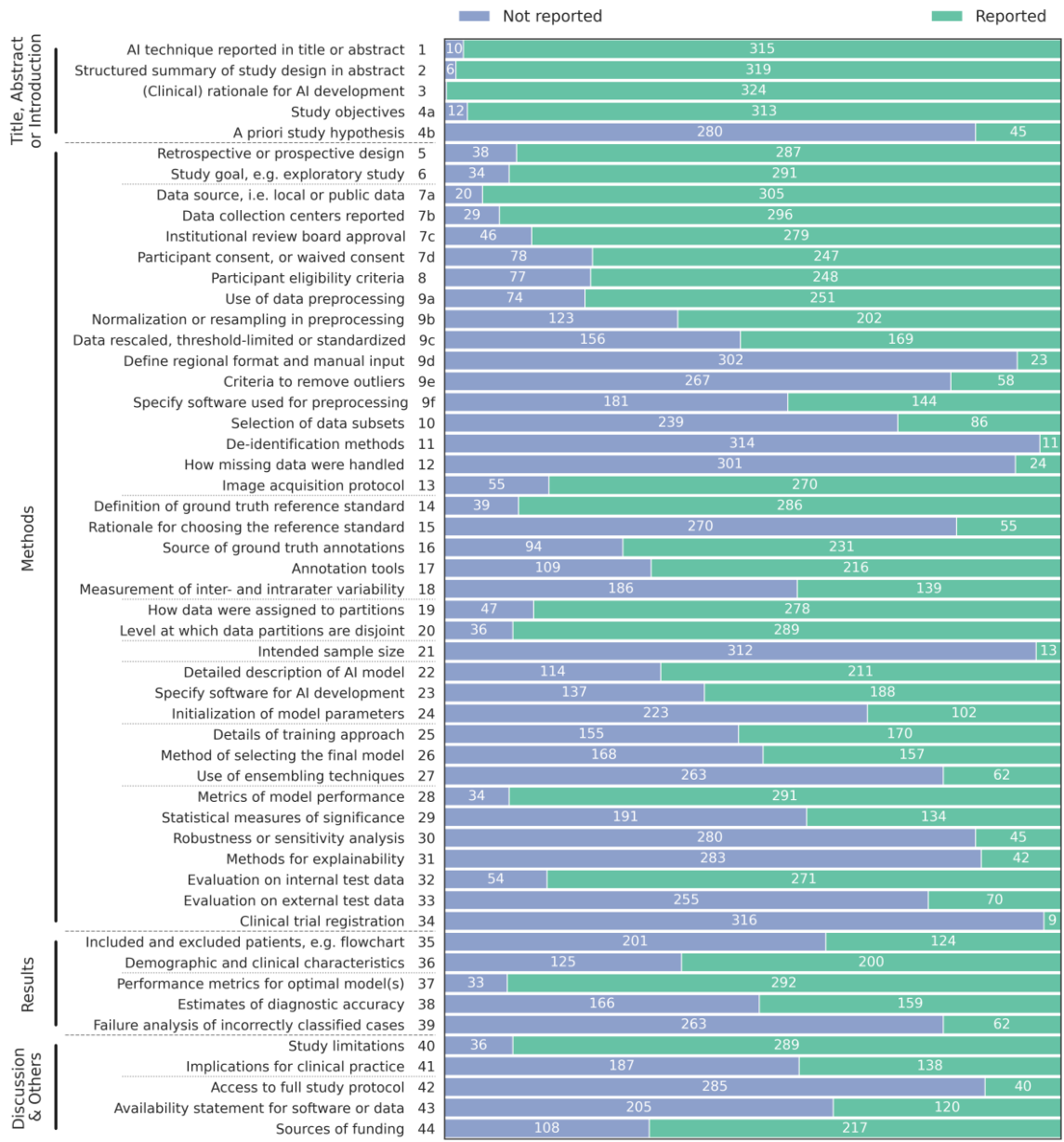

**Figure 5:** Reported and unreported criteria for the included studies (n=325) from the Checklist for Artificial Intelligence in Medical Imaging (CLAIM). Gray bars between criteria within categories indicate subcategories.

**Figure 6:** Scores of the included studies (n=325) for each criterion from the FUTURE-AI international consensus guideline for trustworthy and deployable AI. For each criterion, expected compliance for both research (Res.) and deployable (Dep.) AI tools is reported. F = Fairness, U = Universality, T = Traceability, U = Usability, R = Robustness, E = Explainability, G = General recommendations.

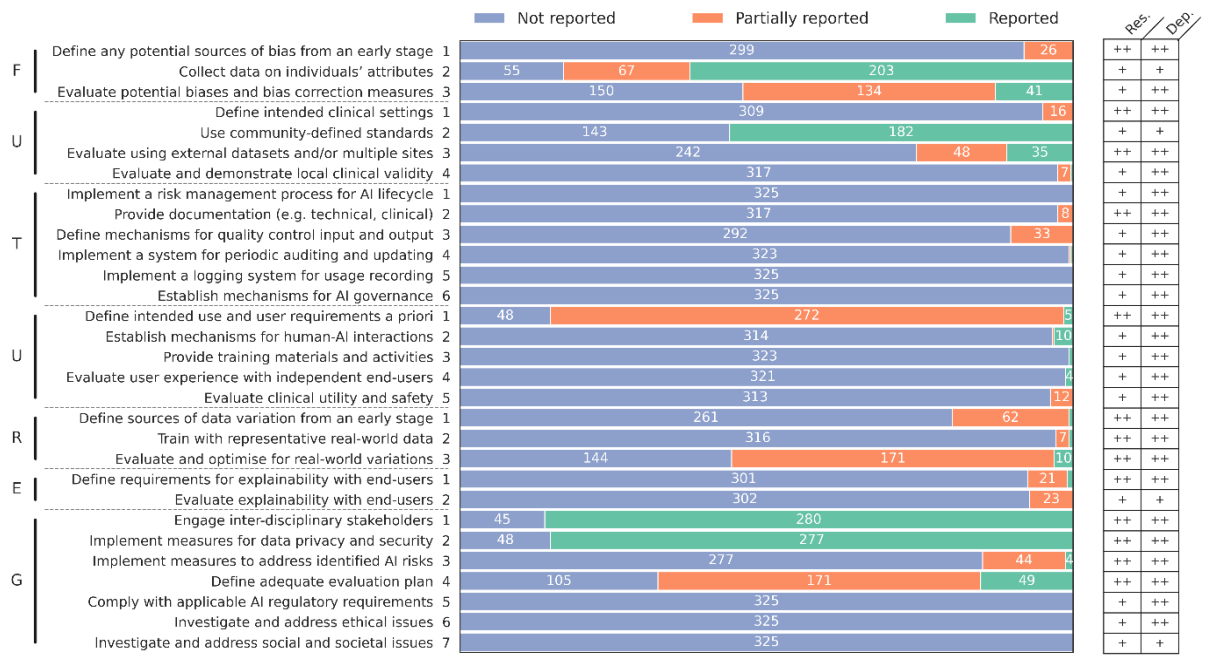

**Supplementary Figures**

**Figure S1:** Inter-reader variability sub-group analysis (n=30) for criteria of the Checklist for Artificial Intelligence in Medical Imaging (CLAIM). Agreement before (green) and after (orange) consensus discussion is reported between raters.

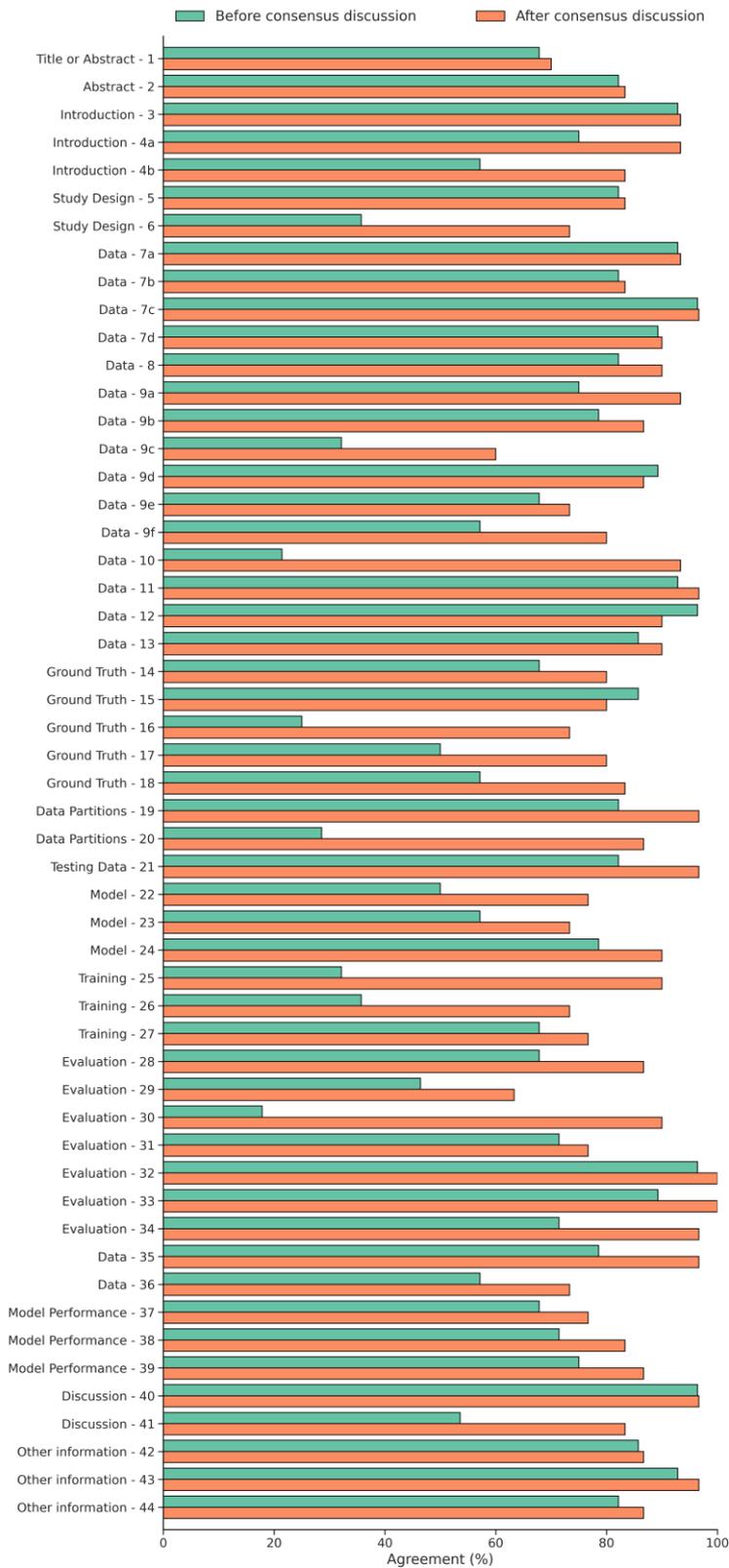

**Figure S2:** Inter-reader variability sub-group analysis (n=30) for criteria of the FUTURE-AI international consensus guideline for trustworthy and deployable AI. Agreement before (green) and after (orange) consensus discussion is reported between raters.

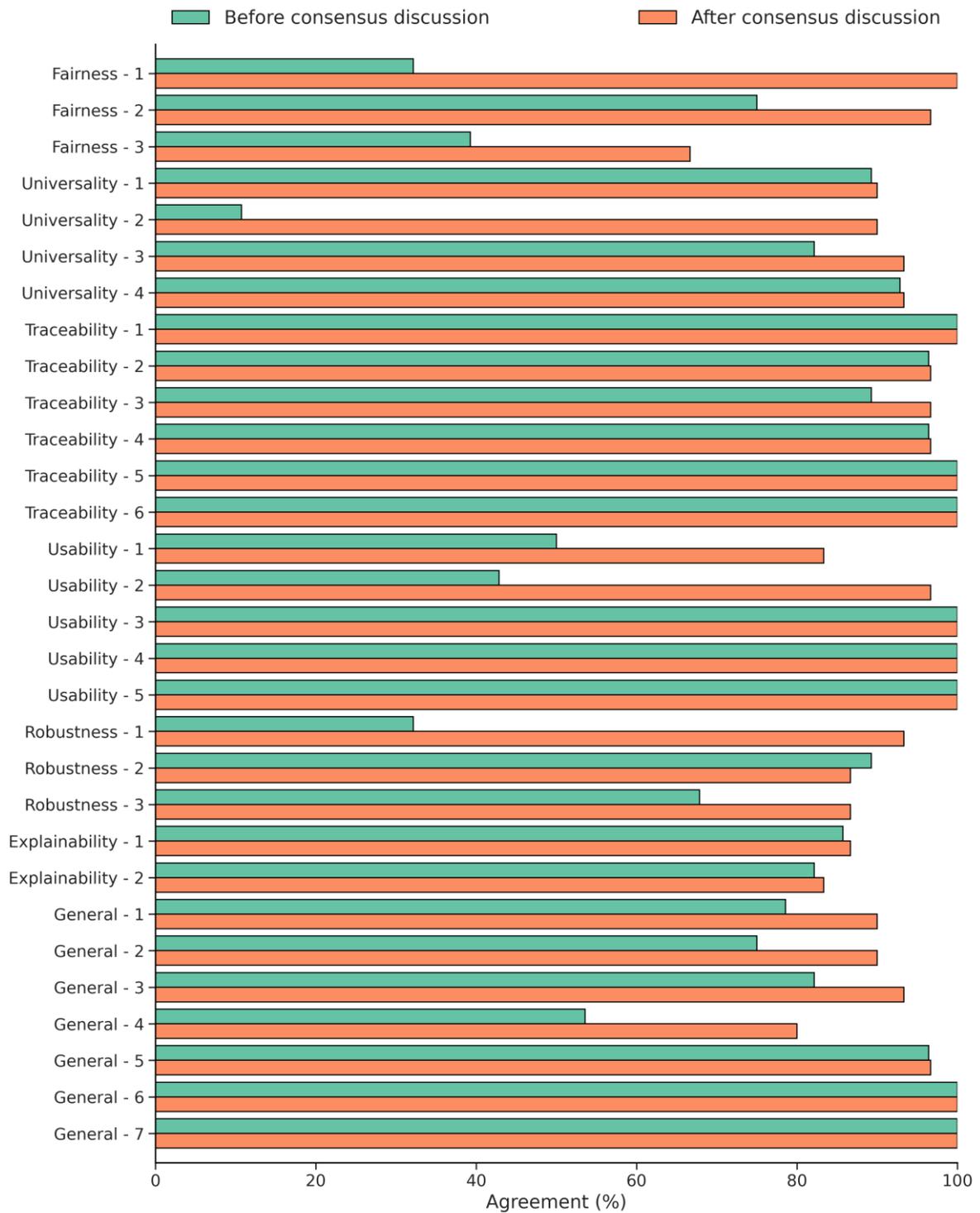

**Figure S3:** Trend of scores on the Checklist for Artificial Intelligence in Medical Imaging (CLAIM) for each year across included studies (n=325). Red dots represent the mean score for each year, while each blue dot corresponds to a single study, with their positions slightly adjusted to avoid overlap. The regression line is calculated with the starting point (x = 0) set to 2008.

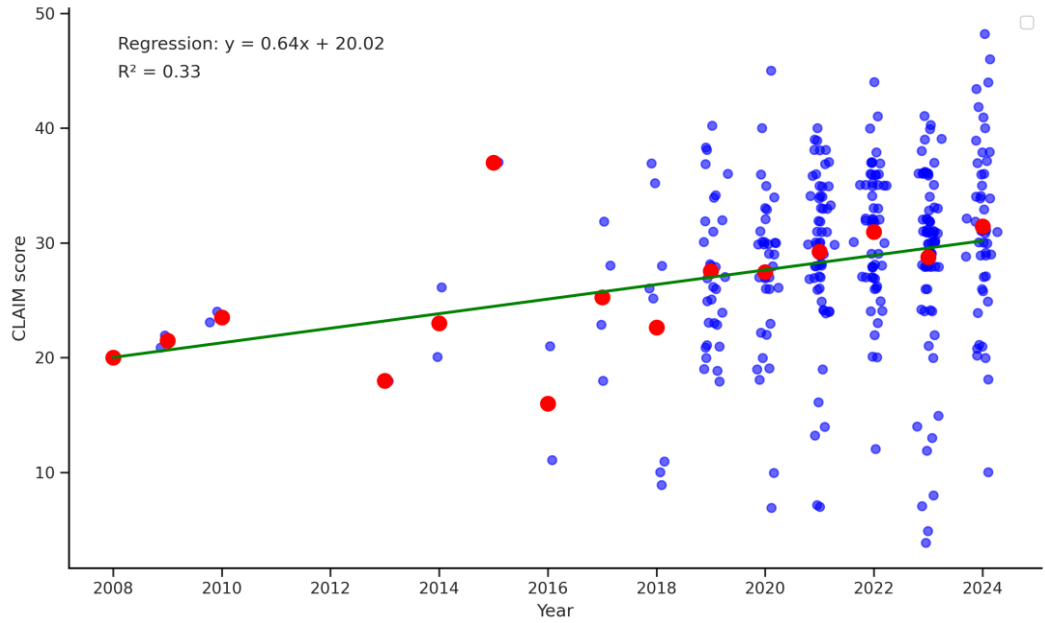

**Figure S4:** Trend of scores on the FUTURE-AI international consensus guideline for trustworthy and deployable AI for each year across included studies (n=325). Red dots represent the mean score for each year, while each blue dot corresponds to a single study, with their positions slightly adjusted to avoid overlap. The regression line is calculated with the starting point (x = 0) set to 2008.

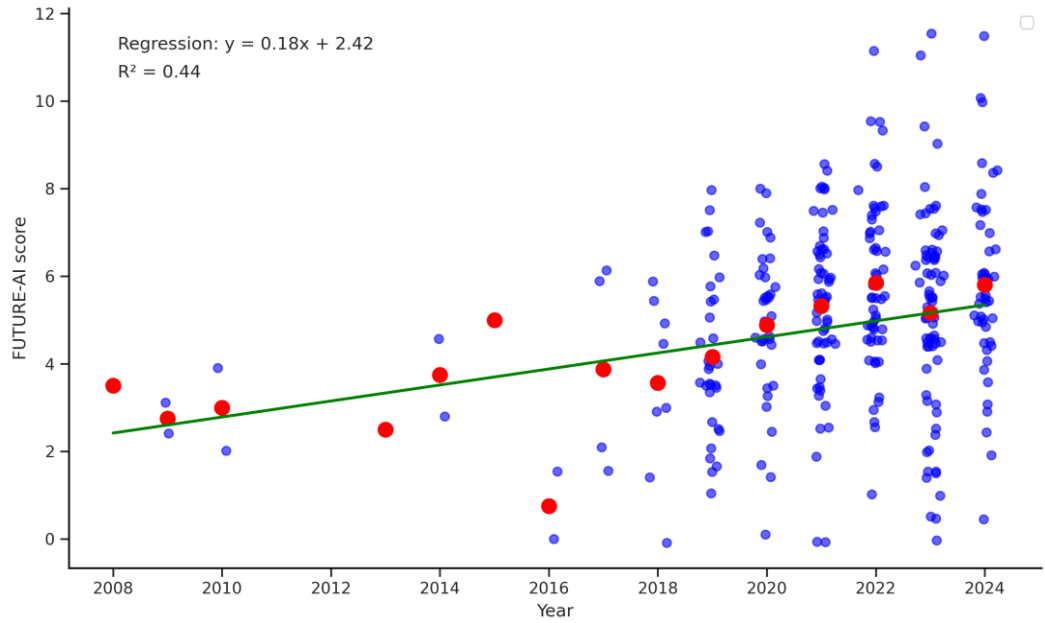

**Figure S5:** Scores on the Checklist for Artificial Intelligence in Medical Imaging (CLAIM) for different AI methods, disease types and predicted outcomes across included studies (n=325).

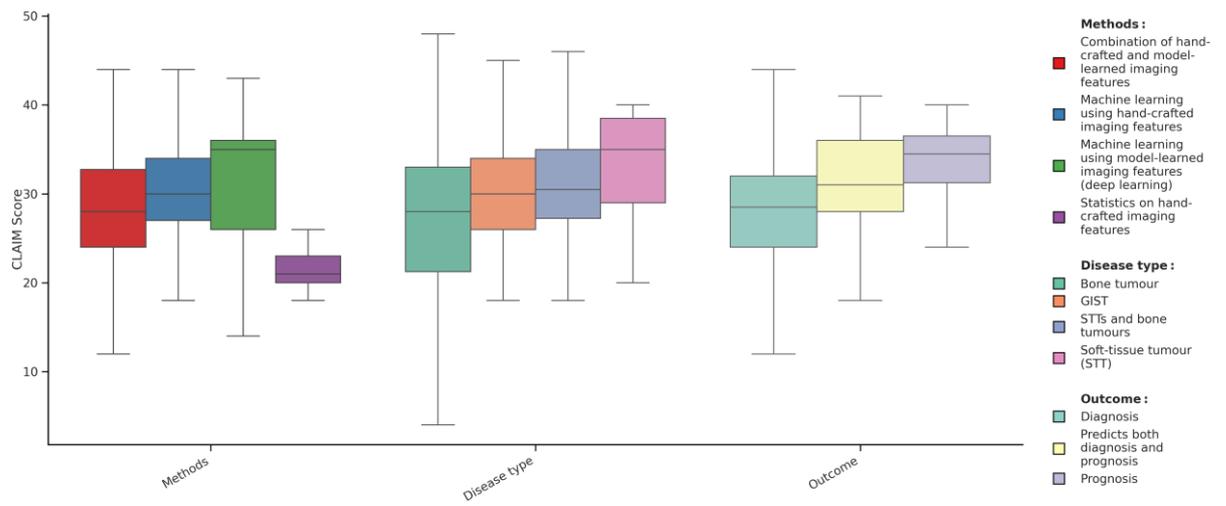

**Figure S6:** Scores on the FUTURE-AI international consensus guideline for trustworthy and deployable AI for different AI methods, disease types and predicted outcomes across included studies (n=325).

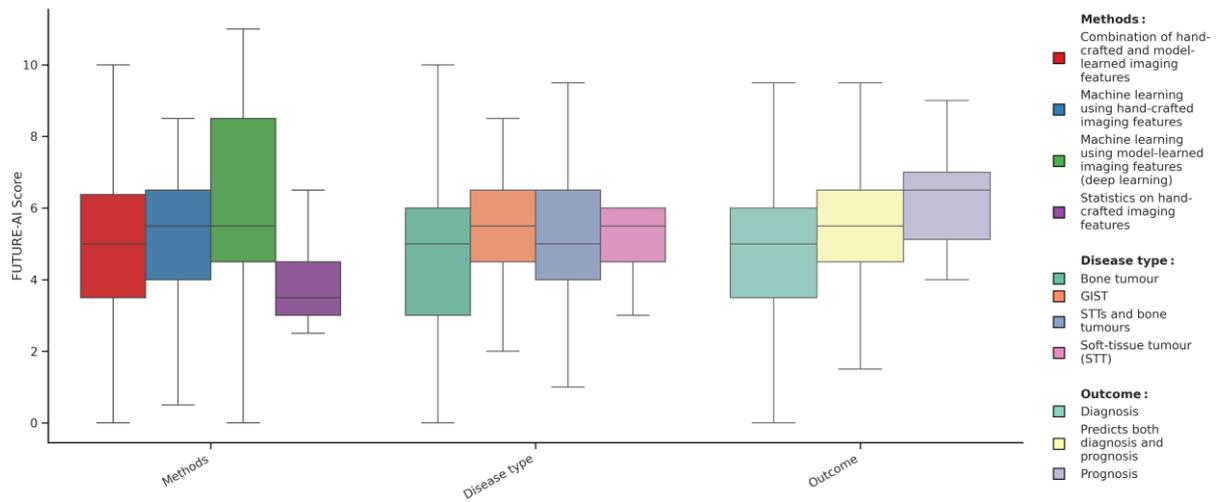

**Figure S7:** Reported and unreported criteria for each study (n=325) from the Checklist for Artificial Intelligence in Medical Imaging (CLAIM). An interactive version of this plot can be found at: [https://douwe-spaanderman.github.io/AI-STTandBoneTumour-Review/](https://douwe-spaanderman.github.io/AI-STTandBoneTumour-Review/)#

**Figure S8:** Scores of each study (n=325) for each criterion from the FUTURE-AI international consensus guideline for trustworthy and deployable AI. An interactive version of this plot can be found at: https://douwe-spaanderman.github.io/AI-STTandBoneTumour-Review/#

**Appendix 1: Search strategy**

| Database searched | Platform | Coverage period |
|---|---|---|
| Medline ALL | Ovid | 1946 – 07/2024 |
| Embase | Embase.com | 1971 - 07/2024 |
| Web of Science Core Collection* | Web of Knowledge | 1975 - 07/2024 |
| Cochrane Central Register of Controlled Trials** | Wiley | 1992 - 07/2024 |
| Additional Search Engines: Google Scholar*** | | |
| Total | | |

*Science Citation Index Expanded (1975- 07/2024); Social Sciences Citation Index (1975- 07/2024); Arts & Humanities Citation Index (1975- 07/2024); Conference Proceedings Citation Index- Science (1990- 07/2024); Conference Proceedings Citation Index- Social Science & Humanities (1990- 07/2024); Emerging Sources Citation Index (2005- 07/2024)

** Manually deleted abstracts from trial registries

***Google Scholar was searched via "Publish or Perish" to download the results in EndNote.

No other database limits were used than those specified in the search strategies

**Embase**

('artificial intelligence'/exp OR 'machine learning'/exp OR 'pattern recognition'/exp OR 'radiomics'/exp OR (CNN OR (artificial* NEAR/3 intelligen*) OR ((machine OR deep) NEAR/3 learning) OR (neural* NEAR/3 network*) OR (classification* NEAR/3 (algorithm OR binary OR multiclass OR multilabel)) OR (classifier*) OR (data-mining*) OR (feature NEAR/3 detection*) OR (feature* NEAR/3 (extraction OR learning OR ranking OR selection OR analysis OR fusion*)) OR (k-nearest* NEAR/3 neighbo*) OR (kernel* NEAR/3 method*) OR (learning* NEAR/3 algorithm*) OR (least* NEAR/3 absolute* NEAR/3 shrinkage* NEAR/3 selection* NEAR/3 operator*) OR (Markov* NEAR/3 model*) OR (memristor*) OR (network* NEAR/3 learning*) OR (perceptron*) OR (radial* NEAR/3 basis* NEAR/3 function*) OR (random* NEAR/3 forest*) OR (recursive* NEAR/3 feature* NEAR/3 elimination*) OR (recursive* NEAR/3 partitioning*) OR (support* NEAR/3 vector* NEAR/3 machine*) OR ((recognition* OR detection* OR classification* OR predict* OR comput* OR diagnos*) NEAR/3 (algorithm* OR network* OR computer-aided* OR automatic* OR automated*)) OR bayesian* OR radiomic* OR pattern-recognit* OR ((AI) NEXT/1 (tool* OR model*))):ab,ti,kw OR AI:ti) **AND** ('musculoskeletal tumor'/exp OR 'bone cyst'/exp OR 'fibrous dysplasia'/exp OR 'lipoma'/exp OR 'hibernoma'/exp OR 'mesenchymoma'/exp OR 'lymphoma'/exp OR 'histiocytosis'/exp OR 'sinus histiocytosis'/exp OR 'sarcoma'/exp OR 'soft tissue tumor'/exp OR 'nerve tumor'/exp OR 'lymphangioma'/exp OR 'lipoblastoma'/exp OR 'ganglion cyst'/exp OR (GCTB OR DDLS OR GIST OR GISTs OR ((soft-tissue* OR adipos*-tissue* OR glomus* OR gastrointest-stroma* OR gastr*-intest*-stroma* OR spinal* OR rib OR skull OR sternal* OR tibial* OR sacrum* OR jaw OR maxillar* OR mandibular* OR odontogenic* OR connective-tissue* OR subcutan*-tissue* OR vein* OR muscle* OR musculoskeletal* OR bone* OR benign-notochordal-cell OR fibrous* OR osteoblast* OR osteoclast* OR synov* OR granular-cell* OR cartilag* OR joint* OR femoral* OR humerus* OR lympho* OR rhabdoid OR non-ossifying OR extramedullary-myeloid* OR atypical-lipmatous* OR nerve* OR giant-cell* OR schwann-cell* OR desmoplastic* OR myofibroblastic*) NEAR/3 (tumor* OR tumour* OR cancer* OR neoplas* OR maligna* OR lesion* OR plasmacytom* OR metasta*)) OR ((vascular* OR arter* OR vessel* OR venal*) NEXT/1 (tumor* OR tumour* OR cancer* OR neoplas* OR maligna* OR lesion* OR plasmacytom* OR metasta*)) OR ((lymph-node*) NEAR/3 (tumor* OR tumour* OR cancer* OR neoplas* OR maligna* OR lesion* OR plasmacytom*)) OR adamantin* OR plasma-cell-granulom* OR glomangiom* OR myoma* OR desmoid* OR Bessel-Hagen OR diaphyseal-aclas* OR ((subungual OR multipl* OR dysplas* OR familial*) NEAR/3 (exosto*)) OR osteocyst* OR ecchondrosis-ossificans OR chondrodysplasia OR adenosarcom* OR sarcom* OR gliosarcoma* OR adenosarcom* OR osteosarcom* OR chondrosarcom* OR chondrom* OR enchondrom* OR chondroblastom* OR chondromatosis* OR osteom* OR osteoblastom* OR osteochondrom* OR maffucci* OR hemangiom* OR haemangiom* OR hemangioendotheliom* OR angiosarcom* OR bone-cyst* OR osseous-cyst* OR intraosseous-gangli* OR intra-osseous-gangli* OR ganglion-cyst* OR jaw-cyst* OR subchondral-cyst* OR chordom* OR synoviom* OR ((fibro*) NEAR/2 (dysplas* OR dystroph* OR osteodys*)) OR cherubism* OR osteofibrous-dysplasi* OR lipom* OR angiolipom* OR angiom* OR lipomatos* OR fetal-lipoma* OR Bannayan OR fatty-kidney OR fatty-pancreas* OR hibernom* OR mesenchym* OR adamantinom* OR hodgkin* OR erdheim-Chester* OR chester-erdheim* OR eosinophil*-granulom* OR histiocytos* OR dorfman-rosai-disease* OR nora-s-lesion* OR chondromesenchymal-hamartoma-of-chest-wall* OR lymphom* OR fibroma* OR osteoclastom* OR histioblastom* OR histiosarcom* OR leiomyosarcom* OR angioendotheliom* OR angioendotheliosarcom* OR hemangiosarcom* OR haemangiosarcom* OR haemangioendotheliom* OR hemangio-endotheliosarcom* OR hemangioendotheliom* OR hemangioendotheliosarcom* OR hemangio-endotheliom OR haemangio-endotheliom* OR lymphangiosarcom* OR Stewart-Treves OR rhabdomysarcom* OR myxofibrosarcom* OR myxosarcom* OR myofibrom* OR myofibroblastom* OR synoviom* OR myxom* OR myopericytom* OR fibrosarcom* OR fibroadenosarcom* OR dermatofibrosarcom* OR neurofibrosarcom* OR chloroma* OR extramedullary-leukaemia* OR extramedullary-leukemia* OR leukosarcom* OR liposarcom* OR neurom* OR perineurom* OR ganglionneurom* OR neurilemom* OR neurofibrom* OR neurothekeom* OR leiomyom* OR rhabdomyom* OR elastofibroma* OR lymphangiom* OR hemangiopericytom* OR haemangiopericytom* OR pericytom* OR myopericytom* OR glomangiopericytom* OR lipoblastom* OR schwannom* OR neurilemmom* OR neurinom* OR neurolemmom* OR neurilemom* OR neurolilemmon* OR ((pigment* OR arthritis*) NEAR/3 (villonodular* OR villous*)) OR ((arthritis*) NEAR/3 (pigment* OR schueller*)) OR ((synovitis*) NEAR/3 (pigment* OR dendritic* OR villonodular*)) OR lymphosarcom* OR reticulosarcom* OR rhabdomyosarcom* OR ameloblastom* OR myosarcom* OR fibrosarcom* OR myoblastom* OR fibrous-histiocytom* OR histiomatos* OR reticulohistiocyt*):ab,ti,kw)
**AND** ('radiomics'/exp OR 'radiogenomics'/exp OR 'diagnostic imaging'/de OR 'radiodiagnosis'/exp OR 'nuclear magnetic resonance imaging'/exp OR 'diffusion coefficient'/de OR 'diffusion weighted imaging'/de OR 'Doppler flowmetry'/de OR 'echography'/exp OR (radiogenomic* OR ((radio OR radiat*) NEXT/1 (genomic* OR diagnos*)) OR radiomic* OR ((diagnos* OR medical*) NEAR/3 imag*) OR radio-genomic* OR radiomic* OR (diagnos* NEAR/3 imag*) OR radiodiagnos* OR ((comput* OR positron) NEAR/3 tomogra*) OR spect OR ct OR pet  OR mri OR (magnetic NEAR/3 resonance) OR ((nuclear OR mr OR multimodalit*) NEAR/3 imaging*) OR rontgen OR roentgen OR ultraso* OR scintigra* OR (diffusion* NEAR/3 (coefficient* OR weighted OR tensor)) OR dwi OR dti OR Doppler OR echogra*):ab,ti,kw) NOT ([Conference Abstract]/lim AND [1800-2020]/py) NOT ('case report'/de OR (case-report):ti) NOT ((animal/exp OR animal*:de OR nonhuman/de) NOT ('human'/exp))

**Medline**
(exp Artificial Intelligence/ OR exp Machine Learning/ OR Pattern Recognition, Automated/ OR (CNN OR (artificial* ADJ3 intelligen*) OR ((machine OR deep) ADJ3  learning) OR (neural* ADJ3 network*) OR (classification* ADJ3 (algorithm OR binary OR multiclass OR multilabel)) OR (classifier*) OR (data-mining*) OR (feature ADJ3 detection*) OR (feature* ADJ3 (extraction OR learning OR ranking OR selection OR analysis OR fusion*)) OR (k-nearest* ADJ3 neighbo*) OR (kernel* ADJ3 method*) OR (learning* ADJ3 algorithm*) OR (least* ADJ3 absolute* ADJ3 shrinkage* ADJ3 selection* ADJ3 operator*) OR (Markov* ADJ3 model*) OR (memristor*) OR (network* ADJ3 learning*) OR (perceptron*) OR (radial* ADJ3 basis* ADJ3 function*) OR (random* ADJ3 forest*) OR (recursive* ADJ3 feature* ADJ3 elimination*) OR (recursive* ADJ3 partitioning*) OR (support* ADJ3 vector* ADJ3 machine*) OR ((recognition* OR detection* OR classification* OR predict* OR comput* OR diagnos*) ADJ3 (algorithm* OR network* OR computer-aided* OR automatic* OR automated*)) OR bayesian* OR radiomic* OR pattern-recognit* OR ((AI) ADJ (tool* OR model*))).ab,ti,kf. OR AI.ti.) **AND** (exp Bone Cysts/ OR exp Fibrous Dysplasia of Bone/ OR exp Lipoma/ OR exp Mesenchymoma/ OR exp Lymphoma/ OR exp Histiocytosis/ OR exp Histiocytosis, Sinus/ OR exp Sarcoma/ OR exp Soft Tissue Neoplasms/ OR exp Neuroma/ OR exp Lymphangioma/ OR exp Ganglion Cysts/ OR (GCTB OR DDLS OR GIST OR GISTs OR ((soft-tissue* OR adipos*-tissue* OR glomus* OR gastrointest-stroma* OR gastr*-intest*-stroma* OR spinal* OR rib OR skull OR sternal* OR tibial* OR sacrum* OR jaw OR maxillar* OR mandibular* OR odontogenic* OR connective-tissue* OR subcutan*-tissue* OR vein* OR muscle* OR musculoskeletal* OR bone* OR benign-notochordal-cell OR fibrous* OR osteoblast* OR osteoclast* OR synov* OR granular-cell* OR cartilag* OR joint* OR femoral* OR humerus* OR lympho* OR rhabdoid OR non-ossifying OR extramedullary-myeloid* OR atypical-lipmatous* OR nerve* OR giant-cell* OR schwann-cell* OR desmoplastic* OR myofibroblastic*) ADJ3 (tumor* OR tumour* OR cancer* OR neoplas* OR maligna* OR lesion* OR plasmacytom* OR metasta*)) OR ((vascular* OR arter* OR vessel* OR venal*) ADJ (tumor* OR tumour* OR cancer* OR neoplas* OR maligna* OR lesion* OR plasmacytom* OR metasta*)) OR ((lymph-node*) ADJ3 (tumor* OR tumour* OR cancer* OR neoplas* OR maligna* OR lesion* OR plasmacytom*)) OR adamantin* OR plasma-cell-granulom* OR glomangiom* OR myoma* OR desmoid* OR Bessel-Hagen OR diaphyseal-aclas* OR ((subungual OR multipl* OR dysplas* OR familial*) ADJ3 (exosto*)) OR osteocyst* OR ecchondrosis-ossificans OR chondrodysplasia OR adenosarcom* OR sarcom* OR gliosarcoma* OR adenosarcom* OR osteosarcom* OR chondrosarcom* OR chondrom* OR enchondrom* OR chondroblastom* OR chondromatosis* OR osteom* OR osteoblastom* OR osteochondrom* OR maffucci* OR hemangiom* OR haemangiom* OR hemangioendotheliom* OR angiosarcom* OR bone-cyst* OR osseous-cyst* OR intraosseous-gangli* OR intra-osseous-gangli* OR ganglion-cyst* OR jaw-cyst* OR subchondral-cyst* OR chordom* OR synoviom* OR ((fibro*) ADJ2 (dysplas* OR dystroph* OR osteodys*)) OR cherubism* OR osteofibrous-dysplasi* OR lipom* OR angiolipom* OR angiom* OR lipomatos* OR fetal-lipoma* OR Bannayan OR fatty-kidney OR fatty-pancreas* OR hibernom* OR mesenchym* OR adamantinom* OR hodgkin* OR erdheim-Chester* OR chester-erdheim* OR eosinophil*-granulom* OR histiocytos* OR dorfman-rosai-disease* OR nora-s-lesion* OR chondromesenchymal-hamartoma-of-chest-wall* OR lymphom* OR fibroma* OR osteoclastom* OR histioblastom* OR histiosarcom* OR leiomyosarcom* OR angioendotheliom* OR angioendotheliosarcom* OR hemangiosarcom* OR haemangiosarcom* OR haemangioendotheliom* OR hemangio-endotheliosarcom* OR hemangioendotheliom* OR hemangioendotheliosarcom* OR hemangio-endotheliom OR haemangio-endotheliom* OR lymphangiosarcom* OR Stewart-Treves OR rhabdomysarcom* OR myxofibrosarcom* OR myxosarcom* OR myofibrom* OR myofibroblastom* OR synoviom* OR myxom* OR myopericytom* OR fibrosarcom* OR fibroadenosarcom* OR dermatofibrosarcom* OR neurofibrosarcom* OR chloroma* OR extramedullary-leukaemia* OR extramedullary-leukemia* OR leukosarcom* OR liposarcom* OR neurom* OR perineurom* OR ganglionneurom* OR neurilemom* OR neurofibrom* OR neurothekeom* OR leiomyom* OR rhabdomyom* OR elastofibroma* OR lymphangiom* OR hemangiopericytom* OR haemangiopericytom* OR pericytom* OR myopericytom* OR glomangiopericytom* OR lipoblastom* OR schwannom* OR neurilemmom* OR neurinom* OR neurolemmom* OR neurilemom* OR neurolilemmon* OR ((pigment* OR arthritis*) ADJ3 (villonodular* OR villous*)) OR ((arthritis*) ADJ3 (pigment* OR schueller*)) OR ((synovitis*) ADJ3 (pigment* OR dendritic* OR villonodular*)) OR lymphosarcom* OR reticulosarcom* OR rhabdomyosarcom* OR ameloblastom* OR myosarcom* OR fibrosarcom* OR myoblastom* OR fibrous-histiocytom* OR histiomatos* OR reticulohistiocyt*).ab,ti,kf.)
**AND** (exp Radiation Genomics/ OR Diagnostic Imaging/ OR exp Magnetic Resonance Imaging/ OR Laser-Doppler Flowmetry/ OR exp Ultrasonography/ OR (radiogenomic* OR ((radio OR radiat*) ADJ1 (genomic* OR diagnos*)) OR radiomic* OR ((diagnos* OR medical*) ADJ3 imag*) OR radiodiagnos* OR ((comput* OR positron) ADJ3 tomogra*) OR spect OR ct OR pet OR mri OR (magnetic ADJ3 resonance) OR ((nuclear OR mr OR multimodalit*) ADJ3 imaging*) OR rontgen OR roentgen OR ultraso* OR scintigra* OR (diffusion* ADJ3 (coefficient* OR weighted OR tensor)) OR dwi OR dti OR Doppler OR echogra*).ab,ti,kf.) NOT (news OR congres* OR abstract* OR book* OR chapter* OR dissertation abstract*).pt. NOT (Case Reports/ OR (case-report).ti.) NOT (exp animals/ NOT humans/)

**Cochrane**
((CNN OR (artificial* NEAR/3 intelligen*) OR ((machine OR deep) NEAR/3 learning) OR (neural* NEAR/3 network*) OR (classification* NEAR/3 (algorithm OR binary OR multiclass OR multilabel)) OR (classifier*) OR (data NEXT/1 mining*) OR (feature NEAR/3 detection*) OR (feature* NEAR/3 (extraction OR learning OR ranking OR selection OR analysis OR fusion*)) OR (k NEXT/1 nearest* NEAR/3 neighbo*) OR (kernel* NEAR/3 method*) OR (learning* NEAR/3 algorithm*) OR (least* NEAR/3 absolute* NEAR/3 shrinkage* NEAR/3 selection* NEAR/3 operator*) OR (Markov* NEAR/3 model*) OR (memristor*) OR (network* NEAR/3 learning*) OR (perceptron*) OR (radial* NEAR/3 basis* NEAR/3 function*) OR (random* NEAR/3 forest*) OR (recursive* NEAR/3 feature* NEAR/3 elimination*) OR (recursive* NEAR/3 partitioning*) OR (support* NEAR/3 vector* NEAR/3 machine*) OR ((recognition* OR detection* OR classification* OR predict* OR comput* OR diagnos*) NEAR/3 (algorithm* OR network* OR computer NEXT/1 aided* OR automatic* OR automated*)) OR bayesian* OR radiomic* OR pattern NEXT/1 recognit* OR ((AI) NEXT/1 (tool* OR model*))):ab,ti,kw OR AI:ti) **AND** ((GCTB OR DDLS OR GIST OR GISTs OR ((soft NEXT/1 tissue* OR adipos* NEXT/1 tissue* OR glomus* OR gastrointest NEXT/1 stroma* OR gastr* NEXT/1 intest* NEXT/1 stroma* OR spinal* OR rib OR skull OR sternal* OR tibial* OR sacrum* OR jaw OR maxillar* OR mandibular* OR odontogenic* OR connective NEXT/1 tissue* OR subcutan* NEXT/1 tissue* OR vein* OR muscle* OR musculoskeletal* OR bone* OR benign NEXT/1 notochordal NEXT/1 cell OR fibrous* OR osteoblast* OR osteoclast* OR synov* OR granular NEXT/1 cell* OR cartilag* OR joint* OR femoral* OR humerus* OR lympho* OR rhabdoid OR non NEXT/1 ossifying OR extramedullary NEXT/1 myeloid* OR atypical NEXT/1 lipmatous* OR nerve* OR giant NEXT/1 cell* OR schwann NEXT/1 cell* OR desmoplastic* OR myofibroblastic*) NEAR/3 (tumor* OR tumour* OR cancer* OR neoplas* OR maligna* OR lesion* OR plasmacytom* OR metasta*)) OR ((vascular* OR arter* OR vessel* OR venal*) NEXT/1 (tumor* OR tumour* OR cancer* OR neoplas* OR maligna* OR lesion* OR plasmacytom* OR metasta*)) OR ((lymph NEXT/1 node*) NEAR/3 (tumor* OR tumour* OR cancer* OR neoplas* OR maligna* OR lesion* OR plasmacytom*)) OR adamantin* OR plasma NEXT/1 cell NEXT/1 granulom* OR glomangiom* OR myoma* OR desmoid* OR Bessel NEXT/1 Hagen OR diaphyseal NEXT/1 aclas* OR ((subungual OR multipl* OR dysplas* OR familial*) NEAR/3 (exosto*)) OR osteocyst* OR ecchondrosis NEXT/1 ossificans OR chondrodysplasia OR adenosarcom* OR sarcom* OR gliosarcoma* OR adenosarcom* OR osteosarcom* OR chondrosarcom* OR chondrom* OR enchondrom* OR chondroblastom* OR chondromatosis* OR osteom* OR osteoblastom* OR osteochondrom* OR maffucci* OR hemangiom* OR haemangiom* OR hemangioendotheliom* OR angiosarcom* OR bone NEXT/1 cyst* OR osseous NEXT/1 cyst* OR intraosseous NEXT/1 gangli* OR intra NEXT/1 osseous NEXT/1 gangli* OR ganglion NEXT/1 cyst* OR jaw NEXT/1 cyst* OR subchondral NEXT/1 cyst* OR chordom* OR synoviom* OR ((fibro*) NEAR/2 (dysplas* OR dystroph* OR osteodys*)) OR cherubism* OR osteofibrous NEXT/1 dysplasi* OR lipom* OR angiolipom* OR angiom* OR lipomatos* OR fetal NEXT/1 lipoma* OR Bannayan OR fatty NEXT/1 kidney OR fatty NEXT/1 pancreas* OR hibernom* OR mesenchym* OR adamantinom* OR hodgkin* OR erdheim NEXT/1 Chester* OR chester NEXT/1 erdheim* OR eosinophil* NEXT/1 granulom* OR histiocytos* OR dorfman NEXT/1 rosai NEXT/1 disease* OR nora NEXT/1 s NEXT/1 lesion* OR chondromesenchymal NEXT/1 hamartoma NEXT/1 of NEXT/1 chest NEXT/1 wall* OR lymphom* OR fibroma* OR osteoclastom* OR histioblastom* OR histiosarcom* OR leiomyosarcom* OR angioendotheliom* OR angioendotheliosarcom* OR hemangiosarcom* OR haemangiosarcom* OR haemangioendotheliom* OR hemangio NEXT/1 endotheliosarcom* OR hemangioendotheliom* OR hemangioendotheliosarcom* OR hemangio NEXT/1 endotheliom OR haemangio NEXT/1 endotheliom* OR lymphangiosarcom* OR Stewart NEXT/1 Treves OR rhabdomysarcom* OR myxofibrosarcom* OR myxosarcom* OR myofibrom* OR myofibroblastom* OR synoviom* OR myxom* OR myopericytom* OR fibrosarcom* OR fibroadenosarcom* OR dermatofibrosarcom* OR neurofibrosarcom* OR chloroma* OR extramedullary NEXT/1 leukaemia* OR extramedullary NEXT/1 leukemia* OR leukosarcom* OR liposarcom* OR neurom* OR perineurom* OR ganglionneurom* OR neurilemom* OR neurofibrom* OR neurothekeom* OR leiomyom* OR rhabdomyom* OR elastofibroma* OR lymphangiom* OR hemangiopericytom* OR haemangiopericytom* OR pericytom* OR myopericytom* OR glomangiopericytom* OR lipoblastom* OR schwannom* OR neurilemmom* OR neurinom* OR neurolemmom* OR neurilemom* OR neurolilemmon* OR ((pigment* OR arthritis*) NEAR/3 (villonodular* OR villous*)) OR ((arthritis*) NEAR/3 (pigment* OR schueller*)) OR ((synovitis*) NEAR/3 (pigment* OR dendritic* OR villonodular*)) OR lymphosarcom* OR reticulosarcom* OR rhabdomyosarcom* OR ameloblastom* OR myosarcom* OR fibrosarcom* OR myoblastom* OR fibrous NEXT/1 histiocytom* OR histiomatos* OR reticulohistiocyt*):ab,ti,kw) **AND** ((radiogenomic* OR ((radio OR radiat*) NEXT/1 (genomic* OR diagnos*)) OR radiomic* OR ((diagnos* OR medical*) NEAR/3 imag*) OR radio NEXT/1 genomic* OR radiomic* OR (diagnos* NEAR/3 imag*) OR radiodiagnos* OR ((comput* OR positron) NEAR/3 tomogra*) OR spect OR ct OR pet OR mri OR (magnetic NEAR/3 resonance) OR ((nuclear OR mr OR multimodalit*) NEAR/3 imaging*) OR rontgen OR roentgen OR ultraso* OR scintigra* OR (diffusion* NEAR/3 (coefficient* OR weighted OR tensor)) OR dwi OR dti OR Doppler OR echogra*):ab,ti,kw) NOT "conference abstract":pt

**Web of Science**
TS=(((CNN OR (artificial* NEAR/2 intelligen*) OR ((machine OR deep) NEAR/2 learning) OR (neural* NEAR/2 network*) OR (classification* NEAR/2 (algorithm OR binary OR multiclass OR multilabel)) OR (classifier*) OR (data-mining*) OR (feature NEAR/2 detection*) OR (feature* NEAR/2 (extraction OR learning OR ranking OR selection OR analysis OR fusion*)) OR (k-nearest* NEAR/2 neighbo*) OR

(kernel* NEAR/2 method*) OR (learning* NEAR/2 algorithm*) OR (least* NEAR/2 absolute* NEAR/2 shrinkage* NEAR/2 selection* NEAR/2 operator*) OR (Markov* NEAR/2 model*) OR (memristor*) OR (network* NEAR/2 learning*) OR (perceptron*) OR (radial* NEAR/2 basis* NEAR/2 function*) OR (random* NEAR/2 forest*) OR (recursive* NEAR/2 feature* NEAR/2 elimination*) OR (recursive* NEAR/2 partitioning*) OR  (support* NEAR/2 vector* NEAR/2 machine*) OR ((recognition* OR detection* OR classification* OR predict* OR comput* OR diagnos*) NEAR/2 (algorithm* OR network* OR computer-aided* OR automatic* OR automated*)) OR bayesian* OR radiomic* OR pattern-recognit* OR ((AI) NEAR/1 (tool* OR model*))) OR AI:ti) AND ((GCTB OR DDLS OR GIST OR GISTs OR ((soft-tissue* OR adipos*-tissue* OR glomus* OR gastrointest-stroma* OR gastr*-intest*-stroma* OR spinal* OR rib OR skull OR sternal* OR tibial* OR sacrum* OR jaw OR maxillar* OR mandibular* OR odontogenic* OR connective-tissue* OR subcutan*-tissue* OR vein* OR muscle* OR musculoskeletal* OR bone* OR benign-notochordal-cell OR fibrous* OR osteoblast* OR osteoclast* OR synov* OR granular-cell* OR cartilag* OR joint* OR femoral* OR humerus* OR lympho* OR rhabdoid OR non-ossifying OR extramedullary-myeloid* OR atypical-lipmatous* OR nerve* OR giant-cell* OR schwann-cell* OR desmoplastic* OR myofibroblastic*) NEAR/2 (tumor* OR tumour* OR cancer* OR neoplas* OR maligna* OR lesion* OR plasmacytom* OR metasta*))  OR ((vascular* OR arter* OR vessel* OR venal*) NEAR/1 (tumor* OR tumour* OR cancer* OR neoplas* OR maligna* OR lesion* OR plasmacytom* OR metasta*)) OR ((lymph-node*) NEAR/2 (tumor* OR tumour* OR cancer* OR neoplas* OR maligna* OR lesion* OR plasmacytom*)) OR adamantin* OR plasma-cell-granulom* OR glomangiom* OR myoma* OR desmoid* OR Bessel-Hagen OR diaphyseal-aclas* OR ((subungual OR multipl* OR dysplas* OR familial*) NEAR/2 (exosto*)) OR osteocyst* OR ecchondrosis-ossificans OR chondrodysplasia OR adenosarcom* OR sarcom* OR gliosarcoma* OR adenosarcom* OR osteosarcom* OR chondrosarcom* OR chondrom* OR enchondrom* OR chondroblastom* OR chondromatosis* OR osteom* OR osteoblastom* OR osteochondrom* OR maffucci* OR hemangiom* OR haemangiom* OR hemangioendotheliom* OR angiosarcom* OR bone-cyst* OR osseous-cyst* OR intraosseous-gangli* OR intra-osseous-gangli* OR ganglion-cyst* OR jaw-cyst* OR subchondral-cyst* OR chordom* OR synoviom* OR ((fibro*) NEAR/2 (dysplas* OR dystroph* OR osteodys*)) OR cherubism* OR osteofibrous-dysplasi* OR lipom* OR angiolipom* OR angiom* OR lipomatos* OR fetal-lipoma* OR Bannayan OR fatty-kidney OR fatty-pancreas* OR hibernom* OR mesenchym* OR adamantinom* OR hodgkin* OR erdheim-Chester* OR chester-erdheim* OR eosinophil*-granulom* OR histiocytos* OR dorfman-rosai-disease* OR nora-s-lesion* OR chondromesenchymal-hamartoma-of-chest-wall* OR lymphom* OR fibroma* OR osteoclastom* OR histioblastom* OR histiosarcom* OR leiomyosarcom* OR angioendotheliom* OR angioendotheliosarcom* OR hemangiosarcom* OR haemangiosarcom* OR haemangioendotheliom* OR hemangio-endotheliosarcom* OR hemangioendotheliom* OR hemangioendotheliosarcom* OR hemangio-endotheliom OR haemangio-endotheliom* OR lymphangiosarcom* OR Stewart-Treves OR rhabdomysarcom* OR myxofibrosarcom* OR myxosarcom* OR myofibrom* OR myofibroblastom* OR synoviom* OR myxom* OR myopericytom* OR fibrosarcom* OR fibroadenosarcom* OR dermatofibrosarcom* OR neurofibrosarcom* OR chloroma* OR extramedullary-leukaemia* OR extramedullary-leukemia* OR leukosarcom* OR liposarcom* OR neurom* OR perineurom* OR ganglionneurom* OR neurilemom* OR neurofibrom* OR neurothekeom* OR leiomyom* OR rhabdomyom* OR elastofibroma* OR lymphangiom* OR hemangiopericytom* OR haemangiopericytom* OR pericytom* OR myopericytom* OR glomangiopericytom* OR lipoblastom* OR schwannom* OR neurilemmom* OR neurinom* OR neurolemmom* OR neurilemom* OR neurolilemmon* OR ((pigment* OR arthritis*) NEAR/2 (villonodular* OR villous*)) OR ((arthritis*)

NEAR/2 (pigment* OR schueller*)) OR ((synovitis*) NEAR/2 (pigment* OR dendritic* OR villonodular*)) OR lymphosarcom* OR reticulosarcom* OR rhabdomyosarcom* OR ameloblastom* OR myosarcom* OR fibrosarcom* OR myoblastom* OR fibrous-histiocytom* OR histiomatos* OR reticulohistiocyt*)) AND ((radiogenomic* OR ((radio OR radiat*) NEAR/1 (genomic* OR diagnos*)) OR radiomic* OR ((diagnos* OR medical*) NEAR/2 imag*) OR radio-genomic* OR radiomic* OR (diagnos* NEAR/2 imag*) OR radiodiagnos* OR ((comput* OR positron) NEAR/2 tomogra*) OR spect OR ct OR pet  OR mri OR (magnetic NEAR/2 resonance) OR ((nuclear OR mr OR multimodalit*) NEAR/2 imaging*) OR rontgen OR roentgen OR ultraso* OR scintigra* OR (diffusion* NEAR/2 (coefficient* OR weighted OR tensor)) OR dwi OR dti OR Doppler OR echogra*)) NOT ((animal* OR rat OR rats OR mouse OR mice OR murine OR dog OR dogs OR canine OR cat OR cats OR feline OR rabbit OR cow OR cows OR bovine OR rodent* OR sheep OR ovine OR pig OR swine OR porcine OR veterinar* OR chick* OR zebrafish* OR baboon* OR nonhuman* OR primate* OR cattle* OR goose OR geese OR duck OR macaque* OR avian* OR bird* OR fish*) NOT (human* OR patient* OR women OR woman OR men OR man))) NOT DT=(Meeting Abstract OR Meeting Summary) NOT TI=(case-report)

**Google Scholar**
"artificial intelligence"|"machine|deep learning"|"neural network"|radiomics "musculoskeletal|bone|nerve tumor|tumour|neoplasm|cancer"|"soft tissue tumor|tumour|neoplasm|cancer" radiomics|radiogenomics|"diagnostic imaging"|"radio diagnosis"|MRI|doppler

# Appendix 2

## WELCOME

This document contains checklists based on the CLAIM [1] and FUTURE-AI [2] guidelines. These checklists were used to assess the quality of research using AI in the diagnosis and prognosis of soft tissue and bone tumours. A completed checklist, used in the study "*AI in radiological imaging of soft-tissue and bone tumours: a systematic review evaluating against CLAIM and FUTURE-AI guidelines* " can be found at: https://douwe-spaanderman.github.io/AI-STTandBoneTumour-Review [3].

The second page in this document (general information) records basic information about each paper and the intial of the reviewer. The third page (FUTURE-AI) gives the checklist based on FUTURE-AI. As well as having a scoring system for each item it is divided into each principle and indiates if an item is "recommended" or "highly-recommended" by the FUTURE-AI guidelines. The fourth page gives the checklist based on CLAIM guidelines. Each item is placed within in its corresponding topic.

The CLAIM checklist has been adapted from the checklist initially developed by Si et al. [4], which used the original version of CLAIM [5] rather than the updated one. The checklist in this document has adapted the checklist created by Si et al. to reflect the 2024 update of CLAIM [1].

| General Information | |
|---|---|
| (sub)Section | Values |
| Rater | |
| Year | |
| Journal | |
| Type imaging | |
| prognosis / diagnosis | 0. Diagnosis 1. Prognosis 2. Both |
| Which disease? | Soft tissue tumour/ Bone tumour/ GIST |
| Used publicly available dataset | 0. No 1. Yes 2. Both |
| Retrospective vs prospective | 0. Retrospective 1. Prospective |
| Signle centre vs multi-centre | 0. Single-centre 1. Multi-centre |
| Data available | 0. No 1. Upon request 2. Yes |
| Code Available | 0. No 1. Upon request 2. Yes |
| Methods | 0. Not learning 1. Hand crafted features 2. Model-learned features 3. Combined |

# FUTURE-AI Checklist

| Principle | no. | Recommendations | Low ML-TRL | High ML-TRL | Description | Scoring criteria |
|---|---|---|---|---|---|---|
| F | 1 | Define any potential sources of bias from an early stage | ++ | ++ | Bias in medical AI is application-specific. At the design phase, the development team should identify possible types and sources of bias for their AI tool. These may include group attributes (e.g. sex, gender, age, ethnicity, socioeconomics, geography), the medical profiles of the individuals (e.g. with comorbidities or disability), as well as human biases (e.g. data labelling, data curation, or the selection of the input features). | 0) No potential biases were discussed prior to AI development; 0.5) Potential biases in at least 1 group (group attributes, medical profile, human biases) were discussed prior to AI development; 1) Potential biases in all 3 groups were discussed prior to AI development. |
| F | 2 | Collect data on individuals' attributes, when possible | + | + | To identify biases and apply measures for increased fairness, relevant attributes of the individuals, such as sex, gender, age, ethnicity, risk factors, comorbidities or disabilities, should be collected. This should be subject to informed consent and approval by ethics committees to ensure an appropriate balance between the benefits for non-discrimination and risks for re-identification. | 0) No relevant attributes of the patient were collected; 0.5) At least the two attributes in the list collected; (list :sex OR gender, age, ethnicity, risk factors(as 1 item), comorbidities or disabilities); 1) More than two attributes in the list were collected, OR with other attributes |
| F | 3 | Evaluate potential biases and bias correction measures | + | ++ | When possible, i.e. the individuals' attributes are included in the data, bias detection methods should be applied by using fairness metrics. To correct for any identified biases, mitigation measures should be applied (e.g. data re-sampling, bias-free representations, equalised odds post-processing) and tested to verify their impact on both the tool's fairness and the model's accuracy. Importantly, any potential bias should be documented and reported to inform the end-users and citizens (see Traceability 2). | 0) Biases were neither investigated nor corrected for; 0.5) Biases were investigated and reported; 1) Biases were also corrected for by mitigation measures (In case of no biases found, 3 also applies) |
| U | 1 | Define intended clinical settings and cross-setting variations | ++ | ++ | At the design phase, the development team should specify the clinical settings in which the AI tool will be applied (e.g. primary healthcare centres, hospitals, home care, low vs. high-resource settings, one or multiple countries), and anticipate potential obstacles to universality (e.g. differences in clinical definitions, medical equipment or IT infrastructures across settings). | 0) The clinical setting was not reported; 0.5) Clinical setting outlined (e.g. primary healthcare centres, hospitals, home care, low vs. high-resource settings, one or multiple countries); 1) Clinical setting outlined and potential obstacles to universality discussed (e.g. differences in clinical definitions, medical equipment or IT infrastructures across settings). |
| U | 2 | Use community-defined standards (e.g. clinical definitions, technical standards) | + | + | To ensure the quality and interoperability of the AI tool, it should be developed based on existing community-defined standards. These may include clinical definitions, medical ontologies (e.g. SNOMED CT, 10 OMOP11), interface standards (e.g. DICOM, FHIR HL7), data annotations, evaluation criteria, and technical standards (e.g. IEEE13 or ISO14). | Are community defined standards used: 0) No 1) Yes |
| U | 3 | Evaluate using external datasets and/or multiple sites | ++ | ++ | To assess generalisability, technical validation of the AI tools should be performed with external datasets that are distinct from those used for training. These may include reference or benchmarking datasets which are representative for the task in question (i.e. approximating the expected real-world variations). Except for AI tools intended for single centres, the clinical evaluation studies should be performed at multiple sites to assess performance and interoperability across clinical workflows. If the tool's generalisability is limited, mitigation measures (e.g. transfer learning or domain adaptation) should be considered, applied and tested. | 0) This study only used single center data --> no external validation; 0.5) Evaluation was performed using external dataset from one other site (or same source, e.g. public available); 1) Evaluation was performed using external dataset from multiple sites; |
| U | 4 | Evaluate and demonstrate local clinical validity | ++ | ++ | Clinical settings vary in many aspects, such as populations, equipment, clinical workflows, and end-users. Hence to ensure trust at each site, the AI tools should be evaluated for their local clinical validity. In particular, the AI tool should fit the local clinical workflows and perform well on the local populations. If the performance is decreased when evaluated locally, re-calibration of the AI model should be performed (e.g., through model fine-tuning or retraining). | 0) local clinical validity has not been discussed, or was not applicable (e.g. AI tool was not deployed outside of research setting/externally); 0.5) local clinical validity has been discussed and evaluated; 1) local clinical validity has been discussed and evaluated and if needed, mitigation strategies have been deployed to deal with this local clinical validity. |
| T | 1 | Implement a risk management process throughout the AI lifecycle | + | ++ | Throughout the AI tool's lifecycle, the development team should analyse potential risks, assess each risk's likelihood, effects and risk-benefit balance, define risk mitigation measures, monitor the risks and mitigations continuously, and maintain a risk management file. The risks may include those explicitly covered by the FUTURE-AI guiding principles (e.g. bias, harm), but also application-specific risks. Other risks to consider include human factors that may lead to misuse of the AI tool (e.g. not following the instructions, receiving insufficient training), application of the AI tool to individuals who are not within the target population, use of the tool by others than the target end-users (e.g. a technician instead of physician), hardware failure, incorrect data annotations or input values, and adversarial attacks. Mitigation measures may include warnings to the users, system shutdown, re-processing of the input data, the acquisition of new input data, or the use of an alternative procedure or human judgement only. | 0) Risks regarding the AI lifecycle have not been described; 0.5) Risks regarding the AI lifecycle have been described; 1) A risk management plan has been described in order to circumvent risks during the AI lifecycle. |
| T | 2 | Provide documentation (e.g. technical, clinical) | ++ | ++ | To increase transparency, traceability, and accountability, adequate documentation should be created and maintained for the AI tool, which may include (i) an AI information leaflet to inform citizens and healthcare professionals about the tool's intended use, risks (e.g. biases) and instructions for use; (ii) a technical document to inform AI developers, health organisations and regulators about the AI model's properties (e.g. hyperparameters), training and testing data, evaluation criteria and results, biases and other limitations, and periodic audits and updates; (iii) a publication based on existing AI reporting standards, and (iv) a risk management file (see Traceability 1). | 0) No documentation has been provided; 0.5) Documentation about 1-2 points (see description) have been provided; 1) Documentation about 3-4 points (see description have been provided) |
| T | 3 | Define mechanisms for quality control of the AI inputs and outputs | + | ++ | The AI tool should be developed and deployed with mechanisms for continuous monitoring and quality control of the AI inputs and outputs, such as to identify missing or out-of-range input variables, inconsistent data formats or units, incorrect annotations or data pre-processing, and erroneous or implausible AI outputs. For quality control of the AI decisions, uncertainty estimates should be provided (and calibrated) to inform the end-users on the degree of confidence in the results. Finally, when necessary, model updates should be applied to address any identified limitations and enhance the AI models over time. | 0) No monitoring or quality control measures of either inputs or outputs have been implemented; 0.5) Monitoring or quality control measures have been implemented for either the inputs or outputs; 1) Monitoring or quality control measures have been implemented for both inputs and outputs. |
| T | 4 | Implement a system for periodic auditing and updating | + | ++ | The AI tool should be developed and deployed with a configurable system for periodic auditing, which should define site-specific datasets and timelines for periodic evaluations (e.g. every year). The periodic auditing should enable the identification of data or concept drifts, newly occurring biases, performance degradation or changes in the decision making of the end-users. Accordingly, necessary updates to the AI models or AI tools should be applied. | 0) No mention of audit or future updating; 0.5) Need of audit or potential updates is discussed; 1) Methods for auditing or updating are discussed. |
| T | 5 | Implement a logging system for usage recording | + | ++ | To increase traceability and accountability, an AI logging system should be implemented to trace the user's main actions in a privacy-preserving manner, including the data that is accessed and used, record the AI predictions and outputs, and log any confounding issues. Time-series statistics and visualisations should be used to inspect the usage of the AI tool over time. | 0) No system has been devised for logging usage of the AI tool; 1) A system has been devised for logging usage of the AI tool |
| T | 6 | Establish mechanisms for AI governance | + | ++ | After deployment, the governance of the AI tool should be specified. In particular, the roles of risk management, periodic auditing, maintenance, and supervision should be assigned, such as to IT teams or healthcare administrators. Furthermore, responsibilities for AI-related errors should be clearly specified among clinicians, healthcare centres, AI developers, and manufacturers. Accountability mechanisms should be established, incorporating both individual and collective liability, alongside compensation and support structures for patients impacted by AI errors. | 0) AI has no governance mechanism (see question for examples); 1) There is at least 1 governance mechanism implemented/described |
| U | 1 | Define intended use and user requirements from an early stage | ++ | ++ | The AI developers should engage clinical experts, end-users (e.g. patients, physicians) and other relevant stakeholders (e.g. data managers, administrators) from an early stage, to compile information on the AI tool's intended use and end-user requirements (e.g. human-AI interfaces), as well as on human factors that may impact the usage of the AI tool (e.g. ergonomics, intuitiveness, experience, learnability). | 0) Only 1 type of stakeholder (e.g. AI developers/departments) was present for AI development, and no intended use and user requirement was described; 0.5) 1 type of stakeholder (e.g. AI developers/departments) was present for AI development, however intended use and end-user requirement was described; OR, multiple stakeholders were present for AI development, no intended use or requirement was described; 1) Multiple stakeholders were present for AI development and compiled information on the AI tool's intended use and end-user requirements. |
| U | 2 | Establish mechanisms for human-AI interactions and oversight | + | + | Based on the user requirements, the AI developers should implement interfaces to enable end-users to effectively utilise the AI model, annotate the input data in a standardised manner, and verify the AI inputs and results. Given the high-stakes nature of medical AI, human oversight is essential and increasingly required by policy makers and regulators. Human-in-the-loop mechanisms should be designed and implemented to perform specific quality checks (e.g. to flag biases, errors or implausible explanations), and to overrule the AI predictions when necessary. | 0) The AI tool has no human oversight; 1) The AI tool provides at least one interface or human-in-the-loop mechanism to involve human oversight |
| U | 3 | Provide training materials and activities (e.g. tutorials, hands-on sessions) | + | + | To facilitate best usage of the AI tool, minimise errors and harm, and increase AI literacy, the developers should provide training materials (e.g. tutorials, manuals, examples) in accessible language and/or training activities (e.g. hands-on sessions), taking into account the diversity of end-users (e.g. clinical specialists, nurses, technicians, citizens or administrators). | Has any training material been provided: 0) No 1) Yes |
| U | 4 | Evaluate user experience and acceptance with independent end-users | + | + | To facilitate adoption, the usability of the AI tool should be evaluated in the real world with representative and diverse end-users (e.g. with respect to sex, gender, age, clinical role, digital proficiency, (dis)ability). The usability tests should gather evidence on the user's satisfaction, performance and productivity. These tests should also verify whether the AI tool impacts the behaviour and decision making of the end-users. | 0) The AI tool was not evaluated for user experience; 0.5) The AI tool was evaluated for user experience by 1 user 1) The AI tool was evaluated for user experience by multiple independent end-users. |
| U | 5 | Evaluate clinical utility and safety (e.g. effectiveness, harm, cost-benefit) | + | + | The AI tool should be evaluated for its clinical utility and safety. The clinical evaluations of the AI tool should show benefits for the clinician (e.g. increased productivity, improved care), for the patient (e.g. earlier diagnosis, better outcomes), and/or for the healthcare organisation (e.g. reduced costs, optimised workflows), when compared to the current standard of care. Additionally, it is important to show that the AI tool is safe and does not cause harm to individuals (or specific groups), such as through a randomised clinical trial. | 0) The AI tool was not evaluated for clinical utility and safety; 0.5) The AI tool was evaluated for clinical utility and safety; 1) The AI tool was evaluated for clinical utility and safety in a Randomized Control Trial (RCT). |
| R | 1 | Define sources of data variation from an early stage | ++ | ++ | At the design phase, an inventory should be made of the application-specific sources of variation that may impact the AI tool's robustness in the real world. These may include differences in equipment, technical fault of a machine, data heterogeneities during data acquisition or annotation, and/or adversarial attacks. | 0) Data acquisition and possible variation of the data source to the real world has not been discussed; 0.5) Data acquisition and possible variation of the data source to the real world has been discussed; 1) Extensive reporting, including reference to the literature and other primary sources, about how the data may vary (or does not vary) to the real world data |
| R | 2 | Train with representative real-world data | ++ | ++ | Clinicians, citizens and other stakeholders are more likely to trust the AI tool if it is trained on data that adequately represents the variations encountered in real-world clinical practice. Hence, the training datasets should be carefully selected, analysed and enriched according to the sources of variation identified at the design phase (see Robustness 1). | 0) The representative of the training data to the real-world data was not evaluated; 0.5) The representative of the training data to the real-world data was evaluated; 1) The representative of the training data to the real-world data was evaluated and enriched accordingly; Note "real world data" has to be data taken from a clinical setting |
| R | 3 | Evaluate and optimise robustness against real-world variations | ++ | ++ | Evaluation studies should be implemented to evaluate the AI tool's robustness (including stress tests and repeatability tests), by considering all potential sources of variation (see Robustness 1), such as data-, equipment-, clinician-, patient- and centre-related variations. Depending on the results, mitigation measures should be implemented to optimise the robustness of the AI model, such as regularisation, data augmentation, data harmonisation, or domain adaptation. | 0) The AI tool has not been evaluated against real-world data (test data); 0.5) The AI tool has been evaluated against real-world data (test data); 1) The AI tool has been evaluated against real-world data (test data) and the AI tool's robustness has been optimized (if applicable) using mitigation methods. |
| E | 1 | Define the need and requirements for explainability with end-users | ++ | ++ | At the design phase, it should be established if explainability is required for the AI tool. In this case, the specific requirements for explainability should be defined with representative experts and end-users, including (i) the goal of the explanations (e.g. global description of the model's behaviour vs. local explanation of each AI decision), (ii) the most suitable approach for AI explainability, and (iii) the potential limitations to anticipate and monitor (e.g. over-reliance of the end-users on the AI potential). | 0) Explainability has not been defined at the design phase; 0.5) At least one of the following areas is discussed: (i) the goal of the explanations (e.g. global description of the model's behaviour vs. local explanation of each AI decision), (ii) the most suitable approach for AI explainability and (iii) the potential limitations to anticipate and monitor (e.g. over-reliance of the end-users on the AI potential). 1) more than one of the areas has been identified and discussed |
| E | 2 | Evaluate explainability with end-users (e.g. correctness, impact on users) | + | + | The explainable AI methods should be evaluated, first quantitatively by using in silico methods to assess the correctness of the explanations, then qualitatively with end-users to assess their impact on user satisfaction, confidence and clinical performance. The evaluations should also identify any limitations of the AI explanations, such as if they are clinically incoherent or sensitive to noise or adversarial attacks, they unreasonably increase the confidence in the AI-generated results. | 0) Explainability has not been defined or not evaluated with end-users; 0.5) Explainability has been evaluated in silico OR with end users involved in the development; 1) Explainability has been evaluated with end-users not involved in the development (e.g. clinical users - radiologists/clinicians, radiographers etc…) |
| General | 1 | Engage inter-disciplinary stakeholders throughout the AI lifecycle | ++ | ++ | Throughout the AI tool's lifecycle, the AI developers should continuously engage with inter-disciplinary stakeholders, such as healthcare professionals, citizens, patient representatives, expert ethicists, data managers and legal experts. This interaction will facilitate the understanding and anticipation of the needs, obstacles and pathways towards acceptance and adoption. | Was a multi-disciplinary team involved in AI development (more than 1 department) 0) No 1) Yes |
| General | 2 | Implement measures for data privacy and security | ++ | ++ | Adequate measures to ensure data privacy and security should be put in place throughout the AI lifecycle. These may include privacy-enhancing techniques (e.g. differential privacy, encryption), data protection impact assessment and appropriate data governance after deployment (e.g. logging system for data access, see Traceability 5). If de-identification is implemented (e.g. pseudonymisation, k-anonymity), the balance between the health benefits for citizens and the risks for re-identification should be carefully assessed and considered. Furthermore, the manufacturers and deployers should implement and regularly evaluate measures for protecting the AI tool against malicious attacks, such as by using system-level cybersecurity solutions or application-specific defence mechanisms (e.g. attack detection or mitigation). | Has data privacy and security been discussed (e.g. data anonymization, clearance from medical ethical committee, (waived) informed consent)? 0) No 1) Yes |
| General | 3 | Implement measures to address identified AI risks | ++ | ++ | At the development stage, the development team should define an AI modelling plan that is aligned with the application-specific requirements. After implementing and testing a baseline AI model, the AI modelling plan should include mitigation measures to address the challenges and risks identified at the design stage (see Fairness 1 to Explainability 1). These may include measures to enhance robustness to real-world variations (e.g. regularisation, data augmentation, data harmonisation, domain adaptation), ensure generalisability across settings (e.g. transfer learning, knowledge distillation), and correct for biases across subgroups (e.g. data re-sampling, bias-free representation, equalised odds post-processing). | 0) No mitigation measures to address challenges and risks identified at the design stage have been reported; 0.5) One mitigation measure (1 of F3, R2 or R3) to either enhance robustness to real-world variation or ensure generalisability across settings or to correct for biases across subgroups has been taken. 1) Two or more mitigation measures (F3, R2, R3) to enhance robustness to real-world variation and/or ensure generalisability across settings and/or correct for biases across subgroups have been taken. |
| General | 4 | Define adequate evaluation plan (e.g. datasets, metrics, reference methods) | ++ | ++ | To increase trust and adoption, an appropriate evaluation plan should be defined (including test data, metrics, and reference methods). First, adequate test data should be selected for assessing each dimension of trustworthy AI. In particular, the test data should be well separated from the training to prevent data leakage. Furthermore, adequate evaluation metrics should be carefully selected, taking into account their benefits and potential flaws. Finally, benchmarking with respect to reference AI tools or standard practice should be performed to enable comparative assessment of model performance. | 0) Evaluation was not conducted using standardized and best practices; 0.5) A separate test set was used to evaluate the AI tool and reported on using appropriate evaluation metrics (e.g. sensitivity and specificity) 1) The AI tool was compared to current standard practice (i.e. evaluation metrics on test set should be compared to same metrics for current clinical tests - so for example how did it compare to radiologists) |
| General | 5 | Identify and comply with applicable AI regulatory requirements | + | ++ | The development team should identify the applicable AI regulations depending on the relevant jurisdictions. This should be done at an early stage to anticipate regulatory obligations based on the medical AI tool's intended classification and risks. | Have AI regulatory requirements been identified? 0) No 1) Yes |
| General | 6 | Investigate and address ethical issues | + | ++ | In addition to the well-known ethical issues that arise in medical AI (e.g. privacy, transparency, equity, autonomy), AI developers, domain specialists and professional ethicists should identify, discuss and address all application-specific ethical, social and societal issues as an integral part of the development and deployment of the AI tool. | Have ethical issues been investigated? 0) No 1) Yes |
| General | 7 | Investigate and address social and societal issues | + | ++ | Social and societal implications should be considered and addressed when developing the AI tool, to ensure a positive impact on citizens and society. Relevant issues include the impact of the AI tool on the working conditions and power relations, on the new skills (or deskilling) of the healthcare professionals and citizens, on future interactions between clinicians, health professionals and social carers. Furthermore, for environmental sustainability, AI developers should consider strategies to reduce the carbon footprint of the AI tool. | Have social and societal implications been investigated? 0) No 1) Yes |

| | | | **CLAIM Checklist** | | |
|---|---|---|---|---|---|
| **(sub)section** | **CLAIM item #** | **Criterion** | | **Explanation** | **Values** |
| Title or Abstract | 1 | The study identifies the AI methodology, or specifies the category of technology used (eg. deep learning). | | Specify the AI techniques used in the study—such as "vision transformers" or "deep learning"—in the article's title and/or abstract; use judgment regarding the level of specificity. | 0. Not specified<br>1. Specified |
| Abstract | 2 | Summary of study design, methods, results, and conclusions | | The abstract should present a succinct structured summary of the study's design, methods, results, and conclusions. Include relevant detail about the study population, such as data source and use of publicly available datasets, number of patients or examinations, number of studies per data source, modalities and relevant series or sequences. Provide information about data partitions and level of data splitting (eg, patient- or image-level). Clearly state if the study is prospective or retrospective and summarize the statistical analysis that was performed. The reader should clearly understand the primary outcomes and implication of the study's findings, including relevant clinical impact. Indicate whether the software, data, and/or resulting model are publicly available (including where to find more details, if applicable). | 0. Not included<br>1. Included |
| Introduction | 3 | Scientific and/or clinical background, including the intended use and role of the AI approach | | **Considered as complete if at least a simple sentence was provided to introduce the medical context and rationale for developing/validating the model**: The current practice should be explicitly mentioned. (1) Describe the study's rationale, goals, and anticipated impact. (2) resent a focused summary of the pertinent literature to describe current practice and highlight how the investigation changes or builds on that work. Guide readers to understand the context for the study, the underlying science, the assumptions underlying the methodology, and the nuances of the study. | 0. Not provided<br>1. Provided |
| | 4a | Study aims and objectives | | Define clearly the clinical or scientific question to be answered; avoid vague statements or descriptions of a process. Limit the chance of post hoc data dredging by specifying the study's hypothesis a priori. The study's hypothesis and objectives should guide appropriate statistical analyses, sample size calculations, and whether the hypothesis will be supported or not. | 0. Not provided<br>1. Provided |
| | 4b | Study hypothesis | | | 0. Not provided<br>1. Provided |
| **Methods** | | | | | |
| Study Design | 5 | Prospective or retrospective study | | Indicate if the study is retrospective or prospective. Evaluate predictive models in a prospective setting, if possible. | 0. Not documented<br>1. Documented |
| | 6 | Study goal | | **Considered as complete if at least a simple sentence was provided involving one of the points below**: (1) Define the study's goal, such as model creation, exploratory study, feasibility study, or noninferiority trial. For classification systems, state the intended use, such as diagnosis, screening, staging, monitoring, surveillance, prediction, or prognosis. (2) Describe the type of predictive modeling to be performed, the target of predictions, and how it will solve the clinical or scientific question. | 0. Not documented<br>1. Documented |
| Data | 7a | Data source | | State the source(s) of data including publicly available datasets and/or synthetic images; provide links to data sources and/or images, if available. Describe how well the data align with the intended use and target population of the model. Provide links to data sources and/or images, if available. Authors are strongly encouraged to deposit data and/or software used for modeling or data analysis in a publicly accessible repository. | 0: Not documented<br>1: Documented |
| | 7b | Data collection institutions | | | 0. Not documented<br>1. Documented |
| | 7c | Institutional review board approval | | | 0. Not documented<br>1. Documented |
| | 7d | Participant consent | | | 0. Not documented<br>1. Documented |
| | 8 | Inclusion and exclusion criteria | | Specify inclusion and exclusion criteria, such as location, dates, patient-care setting, demographics (eg, age, sex, race), pertinent follow-up, and results from prior tests. Define how, where, and when potentially eligible participants or studies were identified. Indicate whether a consecutive, random, or convenience series was selected. | 0. Not provided<br>1. Provided |
| | 9a | Data pre-processing steps with details | | Describe preprocessing steps to allow other investigators to reproduce them. Specify the use of normalization, resampling of image size, change in bit depth, and/or adjustment of window/level settings. If applicable, state whether the data have been rescaled, threshold-limited ("binarized"), and/or standardized. Specify processes used to address regional formatting, manual input, inconsistent data, missing data, incorrect data type, file manipulations, and missing anonymization. State any criteria used to remove outliers. When applicable, include description for libraries, software (including manufacturer name and location and version numbers), and all option and configurations settings. | 0. Not provided<br>1. Provided |
| | 9b | Normalization / resampling in preprocessing | | | 0. Not documented<br>1. Documented |
| | 9c | Whether data have been rescaled, threshold-limited ("binarized"), and/or standardized | | | 0. Not documented<br>1. Documented |
| | 9d | Specify how the following issues were handled: regional format, manual input, inconsistent data, missing data, wrong data types, file manipulations, and missing anonymization. | | | 0. Not documented<br>1. Documented |
| | 9e | Define any criteria to remove outliers | | | 0. Not documented<br>1. Documented |
| | 9f | Specify the libraries, software (including manufacturer name and location), and version numbers, and all option and configuration settings employed. | | | 0. Not documented<br>1. Documented |
| | 10 | Selection of data subsets | | State whether investigators selected subsets of raw extracted data during preprocessing. For example, describe whether investigators selected a subset of the images, cropped portions of images, or extracted segments of a report. If this process is automated, describe the tools and parameters used. If performed manually, describe the training of the personnel and criteria used in their instruction. Justify how this manual step would be accommodated in context of the clinical or scientific problem, describing methods of scaling processes, when applicable. | 0. Not documented<br>1. Documented |
| | 11 | De-identification methods | | Describe the methods used to de-identify data and how protected health information has been removed to meet U.S. (HIPAA), EU (AI Act, EU Health Data Space, GDPR), or other relevant regulations | 0. Not defined<br>1. Defined |
| | 12 | How missing data were handled | | Clearly describe how missing data were handled. For example, describe processes to replace them with approximate, predicted, or proxy values. Discuss biases that imputed data may introduce. | 0. Not defined<br>1. Defined |
| | 13 | Image acquisition protocol | | Describe the image acquisition protocol, such as manufacturer, MRI sequence, ultrasoundfrequency, maximum CT energy, tube current, slice thickness,scan range, and scan resolution; include all relevant parametersto enable reproducibility of the stated methods. | 0. Not defined<br>1. Defined |
| | 14 | Definition of method(s) used to obtain reference standard | | Include a clear, detailed description of methods used to obtain the reference standard; readers should be able to replicate the reference standard based on this description. Include specific, standard guidelines provided to all annotators. Avoid vague descriptions, such as "white matter lesion burden," and use precise definitions, such as "lesion location (periventricular, juxtacortical, infratentorial), size measured in three dimensions, and number of lesions as measured on T2/FLAIR MR brain images." Provide an atlas of examples to annotators to illustrate subjective grading schemes (eg, mild, moderate, severe) and make that information available for review. | 0. Not defined<br>1. Defined |
| | 15 | Rationale for choosing the reference standard | | Describe the rationale for choice of the reference standard versus any alternatives. Include information on potential errors, biases, and limitations of that reference standard. | 0. Not documented<br>1. Documented |

| Category | # | Item | Description | Scoring |
|---|---|---|---|---|
| Ground Truth | 16 | Source of reference standard annotations | Considered as complete if **all points below were provided**: (1) Specify the source of reference standard annotations, citing relevant literature if annotations from existing data resources are used (2) Specify the number of human annotators and their qualifications (eg, level of expertise, subspecialty training). (3) Describe the instructions and training given to annotators; include training materials as a supplement | 0. Not documented<br>1. Documented |
| | 17 | Annotation of test set | Detail the steps taken to annotate the test set with sufficient detail so that another investigator could replicate the annotation. Include any standard instructions provided to annotators for a given task. Specify software used for manual annotation, including the version number. Describe if and how imaging labels were extracted from imaging reports or electronic health records using natural language processing or recurrent neural networks. This information should be included for any step involving manual annotation, in addition to any semiautomated or automated annotation. | 0. Not documented<br>1. Documented |
| | 18 | Measurement of inter- and intrarater variability of features described by annotators | Describe the methods to measure inter- and intra- rater variability, and any steps taken to reduce or mitigate this variability and/or resolve discrepancies between annotators. | 0. Not documented<br>1. Documented |
| Data Partitions | 19 | How data were assigned to partitions; specify proportions | Specify how data were partitioned for training, model optimization (often termed "tuning" or "validation"), and testing. Indicate the proportion of data in each partition (eg, 80/10/10) and justify that selection. Indicate if there are any systematic differences between the data in each partition, and if so, why and how potential class imbalance was addressed. If using openly available data, use established splits to improve comparison to the literature. If freely sharing data, provide data splits so that others can perform model training and testing comparably. | 0. Not documented<br>1. Documented |
| | 20 | Level at which partitions are disjoint | Describe the level at which the partitions are disjoint (eg, patient-, series-, image-level). Sets of medical images generally should be disjoint at the patient level or higher so that images of the same patient do not appear in each partition. | 0. Not documented<br>1. Documented |
| Testing Data | 21 | Intended sample size | Describe the size of the testing set and how it was determined. Use traditional power calculation methods, if applicable, to estimate the required sample size. For classification problems, in cases where there is no algorithm-specific sample size estimation method available, sample size can be estimated for a given area under the curve and confidence interval width | 0. Not documented<br>1. Documented |
| Model | 22 | **Detailed** description of model | If novel model architecture is used, provide a complete and detailed structure of the model, including inputs, outputs, and all intermediate layers, in sufficient detail that another investigator could exactly reconstruct the network. For neural network models, include all details of pooling, normalization, regularization, and activation in the layer descriptions. Model inputs must match the form of the preprocessed data. Model outputs must correspond to the requirements of the stated clinical problem, and for supervised learning should match the form of the reference standard annotations. If a previously published model architecture is employed, cite a reference that meets the preceding standards and fully describe every modification made to the model. Cite a reference for any proprietary model described previously, as well. In some cases, it may be more convenient to provide the structure of the model in code as supplemental data. | 0. Not documented<br>1. Documented |
| | 23 | Software libraries, frameworks, and packages | Specify the names and version numbers of all software libraries, frameworks, and packages. A detailed hardware description may be helpful, especially if computational performance benchmarking is a focus of the work. | 0. Not documented<br>1. Documented |
| | 24 | Initialization of model parameters | Indicate how the parameters of the model were initialized. Describe the distribution from which random values were drawn for randomly initialized parameters. Specify the source of the starting weights if transfer learning is employed to initialize parameters. When there is a combination of random initialization and transfer learning, make it clear which portions of the model were initialized with which strategies. | 0. Not documented<br>1. Documented |
| Training | 25 | Details of training approach | Describe the training procedures and hyperparameters in sufficient detail to enable another investigator to replicate the experiment. To fully document training, a manuscript should: (a) describe how training data were augmented (eg, types and ranges of transformations for images), (b) state how convergence of training of each model was monitored and what the criteria for stopping training were, and (c) indicate the values that were used for every hyperparameter, including which of these were varied between models, over what range, and using what search strategy. For neural networks, descriptions of hyperparameters should include at least the learning rate schedule, optimization algorithm, minibatch size, dropout rates (if any), and regularization parameters (if any). Discuss what objective function was employed, why it was selected, and to what extent it matches the performance required for the clinical or scientific use case. Define criteria used to select the best-performing model. If some model parameters are frozen or restricted from modification, for example in transfer learning, clearly indicate which parameters are involved, the method by which they are restricted, and the portion of the training for which the restriction applies. It may be more concise to describe these details in code in the form of a succinct training script, particularly for neural network models when using a standard framework. | 0. Not documented<br>1. Documented |
| | 26 | Method of selecting the final model | Describe the method and metrics used to select the best-performing model among all the models trained for evaluation against the held-out test set. If more than one model was selected, justify why this was appropriate. | 0. Not documented<br>1. Documented |
| | 27 | Ensembling techniques | If the final algorithm involves an ensemble of models, describe each model comprising the ensemble in complete detail in accordance with the preceding recommendations. Indicate how the outputs of the component models are weighted and/or combined. | 0. Not documented<br>1. Documented |
| Evaluation | 28 | Metrics of model performance | Describe the metrics used to assess the model's performance and indicate how they address the performance characteristics most important to the clinical or scientific problem. Compare the presented model to previously published models. | 0. Not documented<br>1. Documented |
| | 29 | Statistical measures of significance and uncertainty | Considered as complete if **all points below were provided**: (1) Indicate the uncertainty of the performance metrics' values, such as with standard deviation and/or confidence intervals. (2) Compute appropriate tests of statistical significance to compare metrics. (3) Specify the statistical software, including version. | 0. Not documented<br>1. Documented |
| | 30 | Robustness or sensitivity analysis | Analyze the robustness or sensitivity of the model to various assumptions or initial conditions. | 0. Not documented<br>1. Documented |
| | 31 | Methods for explainability or interpretability | If applied, describe the methods that allow one to explain or interpret the model's results and provide the parameters used to generate them. Describe how any such methods were validated in the current study. | 0. Not documented / NA<br>1. Documented |
| | 32 | Evaluation on internal data | Document and describe evaluation performed on internal data. If there are systematic differences in the structure of annotations or data between the training set and the internal test set, explain the differences, and describe the approach taken to accommodate the differences. Document whether there is consistency in performance on the training and internal test sets. | 0. Not described<br>1. Employed internal test data |
| | 33 | Testing on external data | Describe the external data used to evaluate the completed algorithm. If no external testing is performed, note and justify this limitation. If there are differences in structure of annotations or data between the training set and the external testing set, explain the differences, and describe the approach taken to accommodate the differences. | 0. Not described<br>1. Employed external test data |

| Section | # | Item | Description | Scoring |
|---|---|---|---|---|
| | 34 | Clinical trial registration | If applicable, comply with the clinical trial registration statement from the International Committee of Medical Journal Editors (ICMJE). ICMJE recommends that all medical journal editors require registration of clinical trials in a public trials registry at or before the time of first patient enrollment as a condition of consideration for publication. Registration of the study protocol in a clinical trial registry, such as ClinicalTrials.gov or WHO Primary Registries, helps avoid overlapping or redundant studies and allows interested parties to contact the study coordinators. | 0. Not documented<br>1. Documented |
| **Results** | | | | |
| Data | 35 | Flow of participants or cases, using a diagram to indicate inclusion and exclusion | Document the numbers of patients, examinations, or images included and excluded based on each of the study's inclusion and exclusion criteria. Include a flowchart or alternative diagram to show selection of the initial patient population and those excluded for any reason. | 0. Not documented<br>1. Documented |
| Data | 36 | Demographic and clinical characteristics of cases in each partition | Specify the demographic and clinical characteristics of cases in each partition and dataset. Identify sources of potential bias that may originate from differences in demographic or clinical characteristics, such as sex distribution, underrepresented racial or ethnic groups, phenotypic variations, or differences in treatment. | 0. Not documented<br>1. Documented |
| Model Performance | 37 | Performance metrics and measures of statistical uncertainty | Considered as complete if **at least two points below were provided**: (1) Report the final model's performance on the test partition. (2) Benchmark the performance of the AI model against current standards, such as histopathologic identification of disease or a panel of medical experts with an explicit method to resolve disagreements. (3) State the performance metrics on all data partitions and datasets, including any demographic subgroups. | 0. Not documented<br>1. Documented |
| Model Performance | 38 | Estimates of diagnostic accuracy and their precision | Considered as complete if **at least three points below were provided**: For classification tasks, (1) include estimates of diagnostic accuracy and their precision, such as 95% confidence intervals. (2) Apply appropriate methodology such as receiver operating characteristic analysis and/or calibration curves. When the direct calculation of confidence intervals is not possible, report non-parametric estimates from bootstrap samples. (3) State which variables were shown to be predictive of the response variable. (4) Identify the subpopulation(s) for which the prediction model worked most and least effectively. (5) If applicable, recognize the presence of class imbalance (uneven distribution across data classes within or between datasets) and provide appropriate metrics to reflect algorithm performance | 0. Not documented<br>1. Documented |
| Model Performance | 39 | Failure analysis of incorrectly classified cases | Considered as complete if **at least one points below were provided**: Provide information to help understand incorrect results. (1) If the task entails classification into two or more categories, provide a confusion matrix that shows tallies for predicted versus actual categories. (2) Consider presenting examples of incorrectly classified cases to help readers better understand the strengths and limitations of the algorithm. (3) Provide sufficient detail to frame incorrect results in the appropriate medical context. | 0. Not documented<br>1. Documented |
| Discussion | 40 | Study limitations | Identify the study's limitations, including those involving the study's methods, materials, biases, statistical uncertainty, unexpected results, and generalizability. This discussion should follow succinct summarization of the results with appropriate context and explanation of how the current work advances our knowledge and the state of the art. | 0. Not discussed<br>1. Discussed |
| Discussion | 41 | Implications for practice, including the intended use and/or clinical role | Considered as complete if **at least three points below were provided**: (1) Describe the implications for practice, including the intended use and possible clinical role of the AI model. (2) Describe the key impact the work may have on the field. (3) Envision the next steps that one might take to build upon the results. (4) Discuss any issues that would impede successful translation of the model into practice. | 0. Not discussed<br>1. Discussed |
| Other information | 42 | Provide a reference to the full study protocol or to additional technical details | State where readers can access the full study protocol or additional technical details if this description exceeds the journal's word limit. For clinical trials, include reference to the study protocol text referenced in item 34. For experimental or preclinical studies, include reference to details of the AI methodology, if not fully documented in the manuscript or supplemental material. This information can help readers evaluate the validity of the study and can help researchers who want to replicate the study. | 0. Not access to the full study protocol<br>1. Provided access to the full study protocol |
| Other information | 43 | Statement about the availability of software, trained model, and/or data | State where the reader can access the software, model, and/or data associated with the study, including conditions under which these resources can be accessed. Describe the algorithms and software in sufficient detail to allow replication of the study. Authors should deposit all computer code used for modeling and/or data analysis into a publicly accessible repository. | 0. Not discussed<br>1. Discussed |
| Other information | 44 | Sources of funding and other support; role of funders | Specify the sources of funding and other support and the exact role of the funders in performing the study. Indicate whether the authors had independence in each phase of the study. | 0. Not documented<br>1. Documented |

# Appendix 3